\documentclass[pdflatex,sn-mathphys-num]{sn-jnl}

\usepackage{graphicx}%
\usepackage{multirow}%
\usepackage{amsmath,amssymb,amsfonts}%
\usepackage{amsthm}%
\usepackage{mathrsfs}%
\usepackage[title]{appendix}%
\usepackage{xcolor}%
\usepackage{textcomp}%
\usepackage{manyfoot}%
\usepackage{booktabs}%
\usepackage{algorithm}%
\usepackage{algorithmicx}%
\usepackage{algpseudocode}%
\usepackage{listings}
\usepackage{tcolorbox} 
\usepackage{gensymb}

\usepackage{epstopdf}

\theoremstyle{thmstyleone}%
\theoremstyle{thmstyletwo}%

\theoremstyle{thmstylethree}%

\begin{document}

\title[Article Title]{Enhancing Few-Shot Classification of Disaster Imagery with ABHFA-Net}


\author*[1,2]{\fnm{Gao Yu} \sur{Lee}}\email{GAOYU001@e.ntu.edu.sg}
\equalcont{These authors contributed equally to this work.}

\author[2]{\fnm{Tanmoy} \sur{Dam}}\email{tanmoydam@yahoo.com}
\equalcont{These authors contributed equally to this work.}

\author[3]{\fnm{Md.Meftahul} \sur{Ferdaus}}\email{mferdaus@uno.edu}
\equalcont{These authors contributed equally to this work.}

\author[1]{\fnm{Daniel Puiu} \sur{Poenar}}\email{EPDPuiu@ntu.edu.sg}

\author[2]{\fnm{Vu} \sur{Duong}}\email{vu.duong@ntu.edu.sg}

\affil*[1]{\orgdiv{School of Electrical and Electronic Engineering (EEE)}, \orgname{Nanyang Technological University (NTU)}, \orgaddress{\street{50 Nanyang Avenue}, \city{Jurong West}, \postcode{639798}, \state{Singapore}, \country{Singapore}}}

\affil[2]{\orgdiv{School of Mechanical and Aerospace Engineering (MAE)}, \orgname{NTU}, \orgaddress{\street{50 Nanyang Avenue}, \city{Jurong West}, \postcode{639798}, \state{Singapore}, \country{Singapore}}}

\affil[3]{\orgdiv{Department of Computer Science}, \orgname{The University of New Orleans}, \orgaddress{\street{2000 Lakeshore Drive}, \city{New Orleans}, \postcode{70148}, \state{Louisiana}, \country{USA}}}

\abstract{The rising incidence of natural and human-induced disasters necessitates robust visual recognition systems capable of operating under limited labeled data conditions. However, disaster-related image classification remains challenging due to data scarcity, high intra-class variability, and domain-specific complexities in remote sensing imagery. To address these challenges, we propose the Attention Bhattacharyya Distance-based Feature Aggregation Network (ABHFA-Net), a novel few-shot learning (FSL) framework that models class prototypes as probability distributions and performs classification via Bhattacharyya distance-based comparison. Our approach integrates a spatial–channel attention mechanism to enhance discriminative feature learning in the few-shot context and introduces a Bhattacharyya-based contrastive softmax loss for improved class separability. Extensive experiments on both benchmark datasets (CIFAR-FS, FC-100, miniImageNet, tieredImageNet) and real-world disaster datasets (AIDER, CDD, MEDIC) demonstrate the effectiveness of the proposed method. In particular, ABHFA-Net achieves 80.7$\%$ and 92.3$\%$ accuracy on CIFAR-FS under 5-way 1-shot and 5-shot settings, respectively, outperforming existing state-of-the-art methods. On disaster datasets, the model consistently improves classification performance, achieving up to 68.2$\%$ (1-shot) and 78.3$\%$ (5-shot) accuracy on AIDER, highlighting its robustness in real-world scenarios. These results establish ABHFA-Net as a strong and practical solution for few-shot disaster image classification, particularly in data-scarce and time-critical remote sensing applications. The code repository for our work is available at https://github.com/GreedYLearner1146/ABHFA-Net.}

\keywords{Attention Mechanism, Bhattacharyya Distance, Few-Shot Learning, Image Classification, Variational Learning}



\maketitle

\section{Introduction}\label{sec1}


Recent advances in deep learning and computer vision (CV) have substantially improved performance in remote-sensing tasks such as scene classification, segmentation, and object detection. Beyond satellite imagery (e.g., VHR, hyperspectral, SAR) \cite{sun2021research}, low-altitude platforms such as Unmanned Aerial Vehicles (UAVs) now provide high-resolution data that are particularly valuable for disaster monitoring and damage assessment. However, disaster-related remote-sensing datasets are inherently challenging: they are often small-scale, class-imbalanced, and exhibit high intra-class variability due to diverse disaster types, viewpoints, and environmental conditions.

Most state-of-the-art deep learning models rely on large-scale labeled datasets and computationally intensive inference, which limits their applicability to UAV-based or time-critical disaster-response scenarios. Few-Shot Learning (FSL) has emerged as a promising alternative by enabling models to generalize to novel classes using only a handful of labeled samples through meta-learning. The most prevalent approaches in FSL can be categorized as metric-based method, in which a model learn an embedding space where classification is performed via distance-based comparisons between the features extracted from a prior encoder. While these approaches achieve strong performance on benchmark datasets (e.g., miniImageNet \cite{vinyals2016matching}, CIFAR-FS \cite{krizhevsky2009learning}, FC-100 \cite{oreshkin2018tadam}), such datasets are visually distinct from disaster-oriented remote sensing imagery, raising concerns about their transferability to real-world disaster scenarios.


Another fundamental limitation of most existing FSL methods is their reliance on point-estimate class prototypes, typically computed as mean feature embeddings. This formulation is highly sensitive to outliers, limited support samples, and large intra-class variance \cite{zhang2019variational}—conditions that are particularly pronounced in disaster datasets. Variational FSL methods attempt to address this issue by modeling class prototypes as probability distributions rather than deterministic points. However, existing variational FSL approaches primarily rely on Kullback–Leibler (KL) divergence, which measures dissimilarity between distributions but is asymmetric and sensitive to distributional mismatch. Still, there exist few works that approached an alternate distribution comparison metric that are more symmetric, such as the Wasserstein distance \cite{ambrogioni2018wasserstein} and the Hellinger distance \cite{lee2024hela}. Moreover, these variational approaches largely ignore attention mechanisms unlike the point-estimate FSL, which have proven effective in suppressing irrelevant features and enhancing discriminative regions. This is especially true for channel-spatial attention, which even though has been applied widely in CNN-based image classification, had witnessed little applicability in the few-shot context (one work however exists, see \cite{zhang2019channel}) unlike other attention types like cross-attention \cite{hou2019cross}, self-attention \cite{huang2023sapenet}, and attention with weight fusion \cite{meng2023few}. Given that channel-spatial attention is designed to be lightweight and seamless for their incorporation into the learning architecture, it is of interest to explore both its effectiveness and efficiency on the variational FSL, given our strong interest in UAV disaster scenery classification.

In this paper we argued that a variational few-shot inference model that utilized the Bhattacharyya distance \cite{bhattacharyya1946measure} will offer a more principled alternative to address the limitations highlighted above. The Bhattacharyya distance generalizes the Mahalanobis distance \cite{mclachlan1999mahalanobis} and naturally accounts for correlated features, providing a closed-form solution for Gaussian distributions. While such approach has been explored in transfer learning \cite{pandy2022transferability} to quantify class separability, it remains underexplored in few-shot classification, particularly in conjunction with variational modeling and attention mechanisms. To date, no variational FSL framework has jointly investigated distribution-based prototype modeling, attention-enhanced feature aggregation, and Bhattacharyya distance-based classification, especially for disaster-oriented remote sensing datasets.

To this end, we propose the Attention Bhattacharyya Feature Aggregation Net (ABHFA-Net), a novel variational FSL framework tailored for disaster image classification. ABHFA-Net integrates spatial–channel attention into the feature embedding process and models class prototypes as latent Gaussian distributions. Classification is performed by computing the Bhattacharyya distance between query and prototype distributions in the latent space, enabling robust comparison under high variance and limited data. We further derived a modified variational objective—termed $ELBO^{*}$—that establishes a formal connection between Bhattacharyya-based similarity and variational learning. Additionally, we introduced a Bhattacharyya-based contrastive softmax loss to enhance class separability during training. Lastly, we also conducted an ablation study exploring the role of different attention mechanism commonly utilized in FSL in the encoder component of our ABHFA-Net and demonstrated that the channel-spatial attention offered the most efficiency while retaining promising classification performances in the disaster imagery.

In summary, our contributions in this paper are as follows:

\begin{itemize}
   \item We proposed ABHFA-Net, which not only introduced spatial-channel attention in our variational few-shot feature embedding procedure, but also incorporated a Bhattacharyya distance learning approach which serves as a novel VAE strategy in aggregating the class prototypes. To our knowledge, ABHFA-Net is the first of its kind in the domain of FSL which combines channel-spatial attention and the Bhattacharyya distance to perform VAE-based classification of different disaster images in the benchmarked datasets and remote sensing context.
   \item We derived a Bhattacharyya-consistent variational objective ($ELBO^{*}$), establishing a formal mathematical link between the Bhattacharyya coefficient and variational optimization, analogous to existing ELBO formulations in VAE-based models.
   \item We also introduced a Bhattacharyya distance-based contrastive training loss, also known as the Bhattacharyya Softmax loss function $\ell_{BHAS}$, that is inspired by the cosine similarity softmax loss in simCLR \cite{chen2020simple}, but with the cosine similarity in the softmax function replaced by our Bhattacharyya coefficient, since the latter is better suited for class distribution comparisons. The overall training loss for our network, the $\ell_{ABHFA}$, is a weighted linear combination of $\ell_{BHAS}$ and the ordinary categorical cross entropy loss $\ell_{CCE}$. Such loss combinations has been shown to enhance our classification performances relative to the State-Of-The-Arts (SOTA) FSL approaches.
   \item Extensive experiments on four few-shot benchmarked datasets (FC-100, CIFAR-FS, miniImageNet, and tieredImageNet \cite{ren2018meta}), as well as three main disaster-focused datasets—AIDER, CDD, and MEDIC, demonstrate that ABHFA-Net consistently outperforms existing FSL baselines, establishing a strong foundation for few-shot disaster classification in remote-sensing applications. The robustness of our model was further tested on RescueNet and FloodNet, which consist of imagery acquired under more realistic aerial observation conditions, allowing a more realistic simulation of real-time UAV disaster response, as well as testing our model's noisy image/low-resolution robustness and simultaneously extracting efficiency information such as inference latency/Frame-Per-Second (FPS).
\end{itemize}

To emphasize again, although we agreed that attention, variational aggregation, and Bhattacharyya distance are individually established ideas, our ABHFA-Net stands out among existing variational models in their principled integration into a unified distribution-aware few-shot framework. Unlike prior attention-based FSL methods that refine deterministic embeddings, our attention module directly improves the estimation of latent Gaussian prototype parameters. Unlike KL-, Hellinger- and Wasserstein-based variational FSL, ABHFA-Net performs classification through Bhattacharyya overlap between query and class-prototype distributions. Moreover, the proposed $ELBO^{*}$ and $\ell_{BHAS}$ loss make Bhattacharyya similarity part of both the variational objective and discriminative training. Thus, the contribution is not a simple module combination, but a coherent uncertainty-aware prototype learning framework for few-shot disaster image classification.

\section{Related Works}


\subsection{On Few-Shot Learning}

\begin{figure}[hbt!]
    \centering
    \includegraphics[scale=0.78]{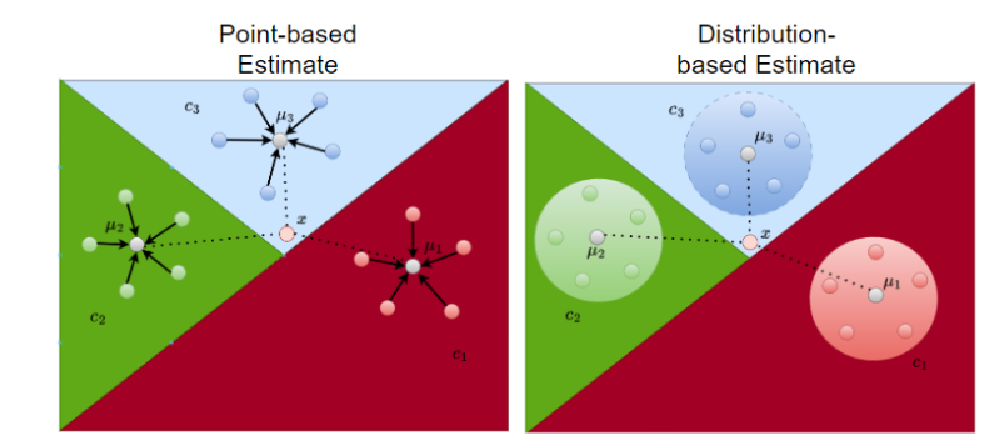}
    \caption{Pictorial depiction of two support feature encoding approaches for a three-class ($c_{1},c_{2},c_{3}$) clustering setting. The grey dot represents the centroid (mean) of each of the class prototype ($\mu_{1},\mu_{2},\mu_{3}$), with its sample points denoted by red, green  and blue respectively. The ``ellipses'' on the right represents the probability distributions of the samples from each distribution. The pale-red dot in the center represents a feature sample point, with the dotted lines denoting its distances to the centroids.}
    \label{fig:VAEvsPts}
\end{figure}

As mentioned in the introductory section, the dominant paradigm in FSL is metric-based learning, where an embedding function is learned such that classification can be performed via distance-based comparisons in the feature space. The latter is depicted in the left of Fig.\ref{fig:VAEvsPts}. Representative approaches include Siamese Networks \cite{koch2015siamese}, Matching Networks \cite{vinyals2016matching}, and Prototypical Networks \cite{snell2017prototypical}, the latter which computes similarities between query and support samples and their corresponding class prototypes. Subsequent works extend this paradigm by learning more expressive similarity functions (e.g., Relation Networks \cite{sung2018learning}) or incorporating transductive inference strategies (e.g., TIM \cite{boudiaf2008transductive}), but they fundamentally still rely on point-estimate feature representations in the embedding space.

Concurrently, FSL has been applied to remote sensing tasks such as classification, semantic segmentation, and object detection. Existing approaches primarily focus on prototype refinement and attention-based feature learning to improve the alignment between support and query samples. For instance, prior works explore dynamic prototype updating (\cite{lang2023global},\cite{cheng2022holistic}), task-adaptive embedding \cite{huang2021taes}, and self-attention \cite{kim2021saffnet} along with multi-scale attention mechanisms \cite{yuan2020few} to enhance discriminative feature extraction.

A key limitation of point-estimate methods is their sensitivity to limited and noisy support samples, as class prototypes are typically computed as simple averages in the embedding space. This issue is particularly pronounced in complex domains such as disaster imagery, where intra-class variability is high and data is scarce. To address this, recent works have explored distribution-based prototype modeling, where each class is represented as a probability distribution rather than a single point \cite{zhang2019variational}. The computation between the class distribution then proceed, which is illustrated at the right of Fig. \ref{fig:VAEvsPts}. For instance, the model by Zhang et al. \cite{zhang2019variational} and VFA-Net \cite{han2023few} leverages variational inference to model class prototypes as latent Gaussian distributions, enabling improved robustness by capturing intra-class variance. Singh et al. \cite{singh2022transductive} utilized TRIDENT which incorporated an attention-based transductive feature extraction module coupled with two decoupled variational inference sub-network. However, existing variational FSL approaches (including the three works described) predominantly rely on Kullback–Leibler (KL) divergence for distribution comparison, which is asymmetric and can be sensitive to distributional mismatch. In disaster-related remote sensing scenarios, the challenges of limited labeled data, class imbalance, and high intra-class variability are further exacerbated. As a result, the support and query samples might have limited overlaps, and the KL divergence used may not be adequate to capture the more subtle underlying data distribution in complex remote sensing images, leading to degraded performance in such FSL context.

\subsection{Distance Measure for Distribution-based Learning}

The choice of distance metric plays a critical role in FSL performance, especially when class representations are modeled as probability distributions. Conventional approaches typically rely on Euclidean or cosine distances, which assume point-based embeddings and do not capture uncertainty or feature correlations. In variational frameworks, KL divergence is widely used due to its connection with the evidence lower bound (ELBO), but it is inherently asymmetric and may be unstable when the support and query distributions have limited overlap.

As mentioned in the introductory section, alternative metrics such as Wasserstein distance and Hellinger distance offer improved theoretical properties, including symmetry and robustness to support mismatch. However, Wasserstein distance often incurs significant computational overhead and lacks a simple closed-form expression under general conditions. The Hellinger distance, while stable and bounded, is typically used as a transformed divergence and does not directly emphasized distributional overlap in a probabilistic sense. In contrast, the Bhattacharyya distance provides a principled measure of similarity by directly quantifying the overlap between two probability distributions. Under Gaussian assumptions, it admits a closed-form expression that incorporates both mean and covariance differences, effectively generalizing Mahalanobis distance. Furthermore, the Bhattacharyya Coefficient ($BC$), for which the corresponding distance is obtained via taking the negative of the logarithm of the coefficient, is related to the square of the Hellinger distance $D^{2}_{H}$ via $BC = 1 - D^{2}_{H}$, allowing us to propose a loss function that closely mimic the Hellinger similarity loss function (as in \cite{lee2024hela}), allowing more stable numerical training and optimization. Despite its desirable properties, the Bhattacharyya distance has been underexplored in FSL, particularly in the context of distribution-based prototype learning and variational modeling, as well as the effect of incorporating various attention mechanisms on such a setting. Our work here aimed to address such gaps and analyzed its performance relative to other distributive-based comparison metric such as the Hellinger and Wasserstein distance, as well as the KL divergence. \\

Feature-fusion-based deep learning approaches have also demonstrated strong performance in limited-data classification settings. Bhosale et al. proposed Thoracic-Net \cite{bhosale2025thoracic}, an explainable AI framework that combines ensemble feature fusion and weighted averaging for few-shot thoracic disease classification from medical images. Their study highlights the importance of robust feature extraction and fusion in improving classification reliability under data scarcity. In contrast to explicit ensemble fusion strategies, our proposed ABHFA-Net performs implicit distribution-aware feature fusion within a variational latent space using Bhattacharyya-distance-based prototype comparison, and applied it to the remote sensing classification instead of a medical setting. \\

\section{Formal Problem Definition}

For the few-shot classification exposition, we posit the existence of a labeled training set designated as $\mathcal{X}_{base} = \{x_{i}, y_{i}\}_{i=1}^{N_{base}}$. Here, the symbol $x_{i}$ represents the raw features pertaining to the $i^{th}$ sample, while $y_{i}$ signifies the one-hot encoded label associated with the same. Within the scope of FSL literature, this labeled ensemble is known as the meta-training (or base) dataset.

Adhering to the FSL paradigm, we consider a test dataset $\mathcal{X}_{test} = \{x_{i}, y_{i}\}_{i=1}^{N_{test}}$ , comprising a novel set of classes represented by $Y_{test}$. The uniqueness of this set is encapsulated in the equation $\mathcal{Y}_{base}\cap \mathcal{Y}_{test} = \emptyset$, illustrating that $\mathcal{Y}_{test}$ introduces completely new class entities not included in $\mathcal{Y}_{base}$. The few-shot scenario involves the creation of randomly sampled few-shot tasks from this test dataset, each encompassing a limited number of labeled instances. In this context,similar $N$-way-$k$-shot task encompasses the random selection of $K_{s}$ labeled instances from each of the $N_{base}$ different classes. The set of these labeled instances, represented by $S$, forms the support set, characterized by its size as $|S| = K_{s} \cdot N_{base}$. Alongside this, each task also possesses a query set denoted as $Q$, consisting of $|Q| = K_{Q} \cdot N_{test}$ unlabeled (or novel) instances from each of the $N_{test}$ classes. The training is reiterated with different image batches presented to the model, and once it is done on the base set, the labeled support sets is utilized for task-specific adaptation. The effectiveness of these techniques is then assessed based on their performance on the unlabeled query sets, providing a measure of the model's capacity to generalize to new data categories based on limited examples.
 
\section{Our Approaches}

In this section, we elaborate on the various key aspects of our ABHFA-Net, as well as the experimental setup and procedures for the evaluation of the FSL SOTAs relative to our proposed design. The loss functions and evaluation metrics utilized throughout our experiments are also highlighted.

\subsection{Components of ABHFA-Net}

There are three novel key components in our ABHFA-Net: the incorporation of channel-spatial attention in the encoder, the Bhattarcharyya distance measure, and the Bhattarcharyya Softmax ($\ell_{BHAS}$) loss function.

\subsubsection{Attention Mechanism}

The attention module we have incorporated into the encoder architecture is composed of channel and spatial attention modules. The proposed attention block is inspired by channel–spatial attention designs such as CBAM \cite{woo2018cbam}, as well as channel-only (SE, ECA) and transformer-based self-attention mechanisms. In our implementation, we adopt a lightweight channel–spatial refinement similar in spirit to CBAM to enhance feature representations before distribution estimation. While the attention module itself is not entirely novel, the key contribution lies in its integration within a variational few-shot learning framework. Specifically, the attention-refined features are used to estimate class-wise latent Gaussian distributions, which are then compared using the Bhattacharyya distance. This differs from prior attention-based few-shot models, where attention is typically used to enhance deterministic embeddings or similarity matching. In contrast, our approach leverages attention to improve the estimation of distribution parameters (mean and covariance), which directly influences probabilistic prototype construction and  distribution-overlap-based classification.

For channel attention, the spatial dimensions of the intermediate feature map $\boldsymbol{\phi}$ are extracted and squeezed using a combination of average and max pooling. The output features from the respective pooling processes are inputted into a Multi-Layer Perceptron (MLP) network, and they are finally combined via an element-wise summation. The aforementioned processes is mathematically described as

\begin{equation}\label{eq1}
\begin{split}
\boldsymbol{M_{c}(\boldsymbol{\phi})} &= \sigma(MLP(AvgPool(\boldsymbol{\phi}) + MLP(MaxPool(\boldsymbol{\phi})))\\
& = \sigma(\boldsymbol{\omega_{1}}(\boldsymbol{\omega_{0}}(\boldsymbol{{\phi^{c}}_{avg}})) +\boldsymbol{\omega_{1}}(\boldsymbol{\omega_{0}}(\boldsymbol{{\phi^{c}}_{max}}))),
\end{split}
\end{equation}

where $\sigma$ is the sigmoid (activation) function, $\boldsymbol{\omega_{0}}$, $\boldsymbol{\omega_{1}}$ are the MLPs weights, $\boldsymbol{M_{c}(\boldsymbol{\phi})}$ is the channel attention map, and $\boldsymbol{{\phi^{c}}_{avg}}$ and $\boldsymbol{{\phi^{c}}_{max}}$ are the feature maps obtained using the aforementioned pooling operations. 

For spatial attention, both aforementioned pooling operations are implemented again, followed by applying a convolutional layer of a filter size of 7 along with the sigmoid activation function to produce the spatial attention map. This is quantitatively described as 

\begin{equation} \label{eq2}
\begin{split}
\boldsymbol{M_{s}(\phi)} &= \sigma(f^{7\times7}([AvgPool(\boldsymbol{\phi}); MaxPool(\boldsymbol{\phi})])) \\
&= \sigma(f^{7\times7}([\boldsymbol{{\phi^{s}}_{avg}}; \boldsymbol{{\phi^{s}}_{max}}])),
\end{split}
\end{equation}

where $\boldsymbol{M_{s}(\phi)}$ denotes the spatial attention map, and $\boldsymbol{{\phi^{s}}_{avg}}$ and $\boldsymbol{{\phi^{s}}_{max}}$ denote the feature maps obtained using the aforementioned pooling operations respectively.

\begin{figure}[hbt!]
    \centering
    \includegraphics[scale=0.5]{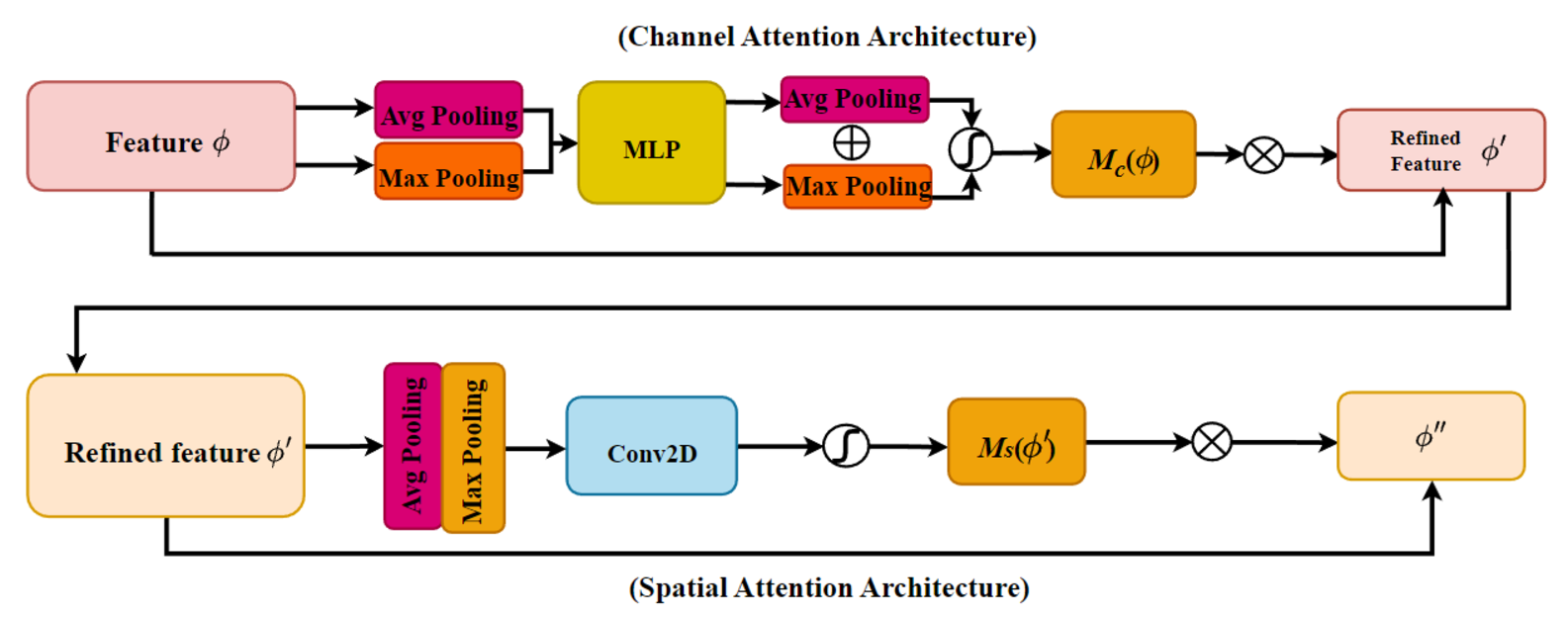}
    \caption{An illustration of the attention model incorporated in the encoder and decoder component of our ABHFA-Net. The architecture comprised of the channel attention and the spatial attention mechanism in which the features are processed consecutively. Note the skip connection between the original and the respective refined features in both the channel and the spatial attention mechanism.}
    \label{fig:AMFA-Net-Attention}
\end{figure}

Altogether, the spatial and channel attention maps are processed as

\begin{equation} \label{eq3}
\begin{split}
\boldsymbol{\phi'} &= \boldsymbol{M_{c}(\phi)} \otimes \boldsymbol{\phi}, \\
\boldsymbol{\phi''} &= \boldsymbol{M_{s}(\phi')} \otimes \boldsymbol{\phi'}, \\
\end{split}
\end{equation}

where $\otimes$ is the tensor product between the extracted features ($\boldsymbol{\phi'}$ and $\boldsymbol{\phi''}$) and the attention maps. The overall architecture of our attention mechanism is illustrated in Fig.\ref{fig:AMFA-Net-Attention}.

\subsubsection{Bhattarcharyya Distance} The Bhattarcharyya distance $D_{BD}$ measure in our ABHFA-Net is applicable to probability distributions acting on the latent variables of the embedded space, and is mathematically depicted as

\begin{equation} \label{eq4}
D_{BD} = -ln \left(\int \sqrt{(p_{\theta}(z|\mathcal{T},\mathcal{S})q_{\phi}(z|\mathcal{S}))}dz \right).
\end{equation}

The integrand with the product term $\sqrt{(p_{\theta}(z|\mathcal{T},\mathcal{S})q_{\phi}(z|\mathcal{S}))}$ is also known as the Bhattarcharyya Coefficient ($BC$). We make the assumption that the prior distribution of $\boldsymbol{z}$, $p_{\theta}(z|\mathcal{T},\mathcal{S})$ \cite{zhang2019variational}, manifested as a multivariate Gaussian $\mathcal{N}(\mu_{i},\sigma_{i})$, whereby the mean is $\mu_{i}$ and the standard deviation is $\sigma_{i}$, and the posterior distribution $q_{\phi}(z|\mathcal{S})$ also took on a similar distribution but with a mean of $\mu_{j}$ and a standard deviation of $\sigma_{j}$ (i.e., $\mathcal{N}(\mu_{j}, \sigma_{j})$). The value of the $\mu$ and $\sigma$ of the support feature set can be extracted from our attention-based encoder network $\mathcal{F}_{encoder}(S)$ after extracting its latent representation $\boldsymbol{z}$ and calculating the distributions. 

In order to make the variation optimization on a similar footing as that of the KL divergence, we can derived a modified Evidence Lower Bound ($ELBO^{*}$) which is related to the Bhattarcharyya distance $D_{BD}$, in a similar manner as that of Lee et al. \cite{lee2024hela}. The mathematical derivation (as well as some of its theoretical properties like convergence analysis and stability proof) are detailed in Appendix A, B and C, but we denote the final mathematical result as



\begin{equation}
\begin{split} \label{eq4.5}
   ELBO^{*} = \int q_{\phi}(z|\mathcal{S})ln\left(p_{\theta}(z|\mathcal{T})\right) dz +2D_{BD},
\end{split}
\end{equation}


where $ELBO^{*}$ can be expressed as $-\int q_{\phi}(z|\mathcal{S})ln\left(\frac{p(z)}{p_{\theta}(z|\mathcal{T}, \mathcal{S})}\right)dz) - ELBO'$, $ELBO' = \int q_{\phi}(z|\mathcal{S})ln\left(\frac{p(z|\mathcal{T},\mathcal{S})}{q_{\phi}(z|\mathcal{S})}\right)dz$, and $\int q_{\phi}(z|\mathcal{S})ln\left(p_{\theta}(z|\mathcal{T})\right) dz$ is the reconstruction term. It can be shown that if $p_{\theta}(z|\mathcal{T}) = q_{\phi}(z|\mathcal{S})$, the $ELBO^{*}$ is equivalent to the evidence term, in agreement with previous related work that associates the respective ELBO with the KL divergence (\cite{zhang2019variational}, \cite{kingma2013auto}), the Wasserstein distance \cite{tolstikhin2017wasserstein} and the Hellinger distance \cite{lee2024hela}.

\subsubsection{Bhattarcharyya Softmax (BHAS) Loss} Our proposed new training loss function also consisted of BHAS loss $\ell_{BHAS}$. As mentioned in the introductory section, the modification made here is that instead of the cosine similarity, the Bhattarcharyya coefficient is directly incorporated into the parentheses of the softmax formula as highlighted below: 

\begin{equation} \label{eq5}
\ell_{BHAS} = -log\left(\frac{e^{(\frac{BC}{\tau})}}{\sum_{i=1}^{2N}e^{(\frac{BC}{\tau})}}\right),
\end{equation}

where $\tau$ denotes the temperature parameter of the softmax function, set as 0.01 throughout our training. Here, we omitted the notation of $q_{\phi}(z|S)$ and $p_{\theta}(z|\mathcal{T}, \mathcal{S})$ in $BC$ and implicitly implied that the coefficient is a function of the two distribution. However, during our simulation, we have found that utilizing the loss function in the mathematical form as above would lead to gradient explosion due to the instability incurred when multiplying two probability distributions together, in accord to the definition of $BC$. To address this problem, we recalled that the Bhattarcharyya coefficient is related to the Hellinger distance $D_{H}$ \cite{hellinger1909neue} as $D_{H} = \sqrt{1 - \int\sqrt{(p_{\theta}(z|\mathcal{T}, \mathcal{S})q_{\phi}(z|\mathcal{S}))}dz} = \sqrt{1 - BC}$. Therefore, by rewriting the Bhattarcharyya coefficient in terms of the squared of the Hellinger distance $D_{H}^{2}$ as $BC = 1 - D_{H}^{2}$, our $\ell_{BHAS}$ can be re-written as 


\begin{equation} \label{eq5.5}
\ell_{BHAS} = -log\left(\frac{e^{\left(\frac{1 - D_{H}^{2}}{\tau}\right)}}{\sum_{i=1}^{2N} e^{\left(\frac{1 - D_{H}^{2}}{\tau}\right)}}\right),
\end{equation}

where in the above mathematical form, the gradient explosion disappear as the Hellinger distance is defined as the sum of the square of the differences in the square root of each distribution $D_{H} = \int (\sqrt{p} - \sqrt{q})^{2} dz$, which implies that the gradient computation can be performed in a more stable manner. Note that our $\ell_{BHAS}$ closely resembled that of the ordinary negative log likelihood loss function $\ell_{nll}$ used for multi-class classification. The motivation of such loss is that it generalizes the softmax loss in simCLR to address distribution comparison. Since the cosine similarity was used in the former, the underlying assumption is that the embedded features are of point-estimate, and they also could not account for possible collinear probabilistic distribution configurations that hint some level of correlations among the features. Similarly to softmax loss in simCLR, in the denominator, we summed the numbers in parentheses over the number of positive samples and divided by two times the batch size $N$ selected per training epoch (i.e., $2N$). Like the Bhattarcharyya distance, some theoretical properties like stability proofs of $\ell_{BHAS}$ are detailed in Appendix C.

\begin{figure*}[t]        
    \centering   
    \includegraphics[width=1.2\linewidth]{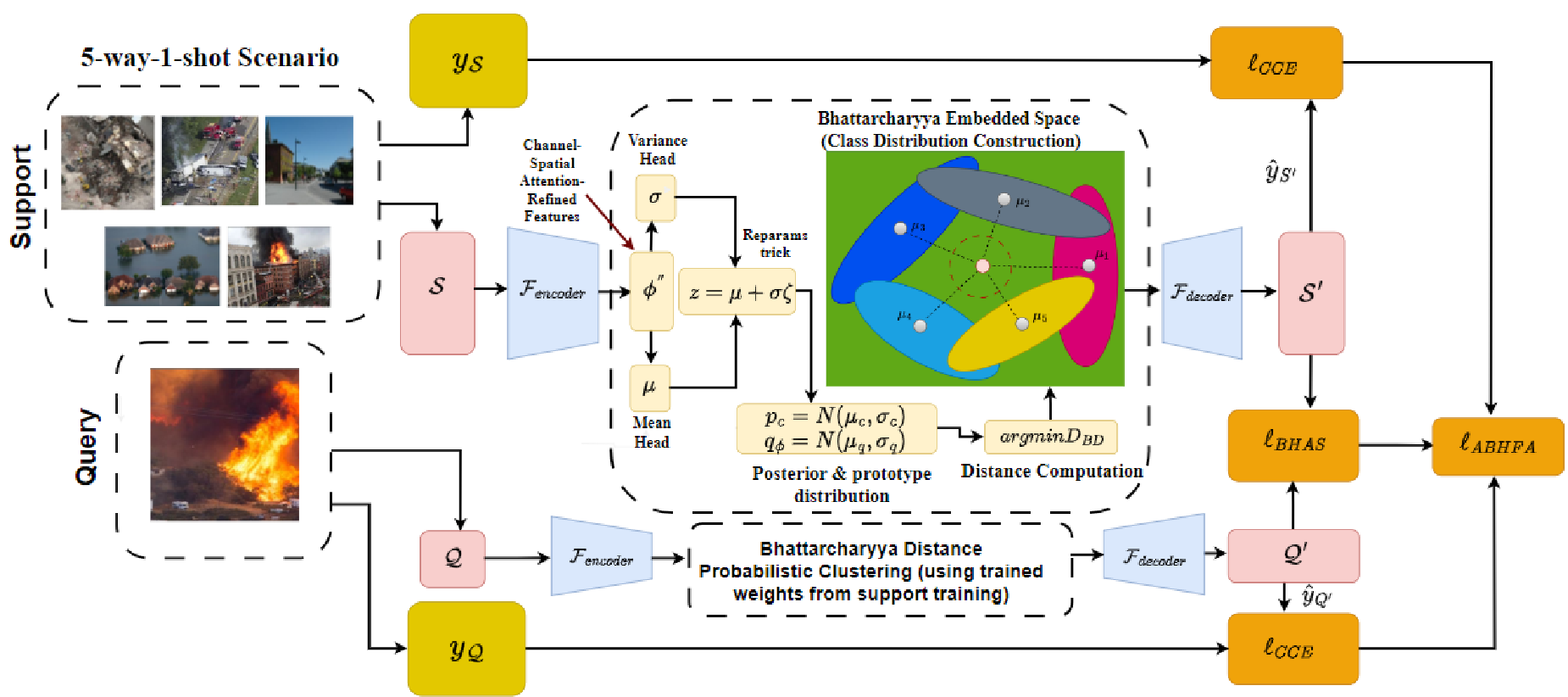}
    \caption{The illustration of our ABHFA-Net architecture. It receives a set of support and query images as input, which are then processed by an attention-based encoder to extract their features. These features undergo encoding into latent representation in which the respective Gaussian probability distributions are obtained. The Bhattarcharyya distance computation process involves deriving confidence spaces for each class prototype's distribution and comparing their overlapping degree with feature point distribution (represented by a red dot within a red dotted circle). The classes' maximum probabilities are leveraged to generate a confidence score, indicating the predicted label for a given sample. To perform training via the Bhattacharyya Softmax loss $\ell_{BHAS}$ and categorical cross-entropy loss $\ell_{CCE}$ using the resultant support and query sets ($S'$, $Q'$), we use ABHFA-Net loss: $\ell_{ABHFA}$. $y_{\mathcal{S}}$ and $y_{\mathcal{Q}}$ denote the ground-truth label from the support and query set image respectively, and $\hat{y}_{\mathcal{S}}$ and $\hat{y}_{\mathcal{Q}}$ denote the predicted label from the support and query set image respectively. The four quantities are relevant for computing the $\ell_{CCE}$. The diagram presented above illustrates a 5-way-1-shot Fast FSL training method.}
    \label{fig:AMFA-Net}
\end{figure*}


\begin{algorithm}[t]
\caption{ABHFA-Net Episodic Training and Inference}
\label{alg:abhfa}
\begin{algorithmic}[1]

\Require Meta-training dataset $\mathcal{X}_{base}$, number of classes per episode $N_c$, number of support samples $N_S$, number of query samples $N_Q$
\Ensure Predicted labels $\hat{y}$ for query set

\For{each training episode}
    \State Randomly sample $N_c$ classes $\mathcal{C}_{subset}$ from $\mathcal{X}_{base}$
    
    \For{each class $c \in \mathcal{C}_{subset}$}
        \State Randomly sample $N_S$ support samples $S_c$
        \State Randomly sample $N_Q$ query samples $Q_c$, where $S_c \cap Q_c = \emptyset$
    \EndFor
    
    \State Construct support set $S = \bigcup_c S_c$ and query set $Q = \bigcup_c Q_c$
    
    \State \textbf{Feature Encoding:} Encode $S$ and $Q$ using attention-based encoder $F_{\text{encoder}}$
    
    \State \textbf{Latent Representation:} Obtain latent variables $z = \mu + \zeta\sigma$, where $\zeta \sim \mathcal{N}(0,I)$
    
    \For{each class $c$}
        \State Estimate class prototype distribution $p_c = \mathcal{N}(\mu_c,\sigma_c)$ from $S_c$
    \EndFor
    
    \For{each query sample $q \in Q$}
        \State Compute query distribution $q_{\phi}(z|Q)$
        
        \For{each class $c$}
            \State Compute $D_{BD}(q,p_c)$ using Eq.~\ref{eq4}
        \EndFor
        
        \State Predict label $\hat{y} = \arg\min_c D_{BD}(q,p_c)$
    \EndFor
    
    \State Compute $\ell_{\text{BHAS}}$ using Eqs.~\ref{eq5} and~\ref{eq5.5}, and $\ell_{\text{CCE}}$ using Eq.~\ref{eq7}
    
    \State Compute total loss $\ell_{\text{ABHFA}} = \lambda_1 \ell_{\text{BHAS}} + \lambda_2 \ell_{\text{CCE}}$
    
    \State Update model parameters via backpropagation
\EndFor

\end{algorithmic}
\end{algorithm}


\subsection{ABHFA-Net Architecture} Altogether, our ABHFA-Net architecture is illustrated in Fig.\ref{fig:AMFA-Net}, which also clearly depicts the aforementioned components. For clarity, we describe the forward pass of ABHFA-Net step-by-step. Given a support set $S$ and query set $Q$, both are first passed through a shared encoder $F_{\text{encoder}}$, where channel-spatial attention is applied to produce refined feature representations $\boldsymbol{\phi''}$. These features are then projected into latent distribution parameters, where two parallel heads estimate the mean $\mu$ and standard deviation $\sigma$. Using the reparameterization trick, latent variables are sampled as $z = \mu + \zeta \sigma$, where $\zeta \sim \mathcal{N}(0, I)$. For each class $c$, support samples are grouped to construct class prototype distributions $p_{\theta} = \mathcal{N}(\mu_c, \sigma_c)$. Similarly, each query sample is represented as a distribution $q_{\phi} = \mathcal{N}(\mu_q,\sigma_q)$. Classification is performed by computing the Bhattacharyya distance between the query distributions and each class prototype distribution.

The previous step comprises the comparison of the selected feature distribution in the embedding space (denoted by a red cross surrounded by the red dotted circle) in Fig.\ref{fig:AMFA-Net}) with that of the prototypes' feature points by measuring the degree of overlap between the respective distributions. These series of steps closely mimicked that of VFA-Net except with the addition of attention in the encoder and our Bhattarcharyya probability comparison structure. The respective class-specific distributions that are estimated from the sets are utilized to calculate the probability of the target data, which is the maximum probabilistic attainable among the class. This is then used to generate a confidence score for the label to which the sample is most likely to belong (predicted label), in agreement with \cite{zhang2019variational}. Throughout the above processes, the resultant support $S'$ and the query's probabilistic features $Q'$ are guided by the loss functions $\ell_{BHAS}$ and $\ell_{CCE}$, which are combined for the network training on the sets. Therefore, the total loss for our model training $\ell_{ABHFA}$ is described by

\begin{equation} \label{eq6}
\ell_{ABHFA} = \lambda_{1} \cdot \ell_{BHAS} + \lambda_{2} \cdot \ell_{CCE},
\end{equation}

in which $\ell_{CCE}$ is mathematically described by

\begin{equation} \label{eq7}
\begin{split}
\ell_{CCE} = -\sum_{i}^{Q} y_{i}log \left(p(\hat{y}_{i} = y_{i}|\mathcal{T}_{meta})\right),
\end{split}
\end{equation}

where $y_{i}$ and $\hat{y}_{i}$ represents the ground-truth and predicted labels, respectively, $\mathcal{T}_{meta}$ denotes a meta-task and runs through the total number of queries $Q$ per training batch in our experiment. The $\lambda_{1}$ and $\lambda_{2}$ are hyperparameters which are set to 0.5 and 1.0 respectively. The algorithmic outline of our ABHFA-Net is described in Algorithm 1.

\section{Experiments}

\begin{figure}[hbt!]
    \centering
    \includegraphics[height = 8.5cm, width=11.05cm]{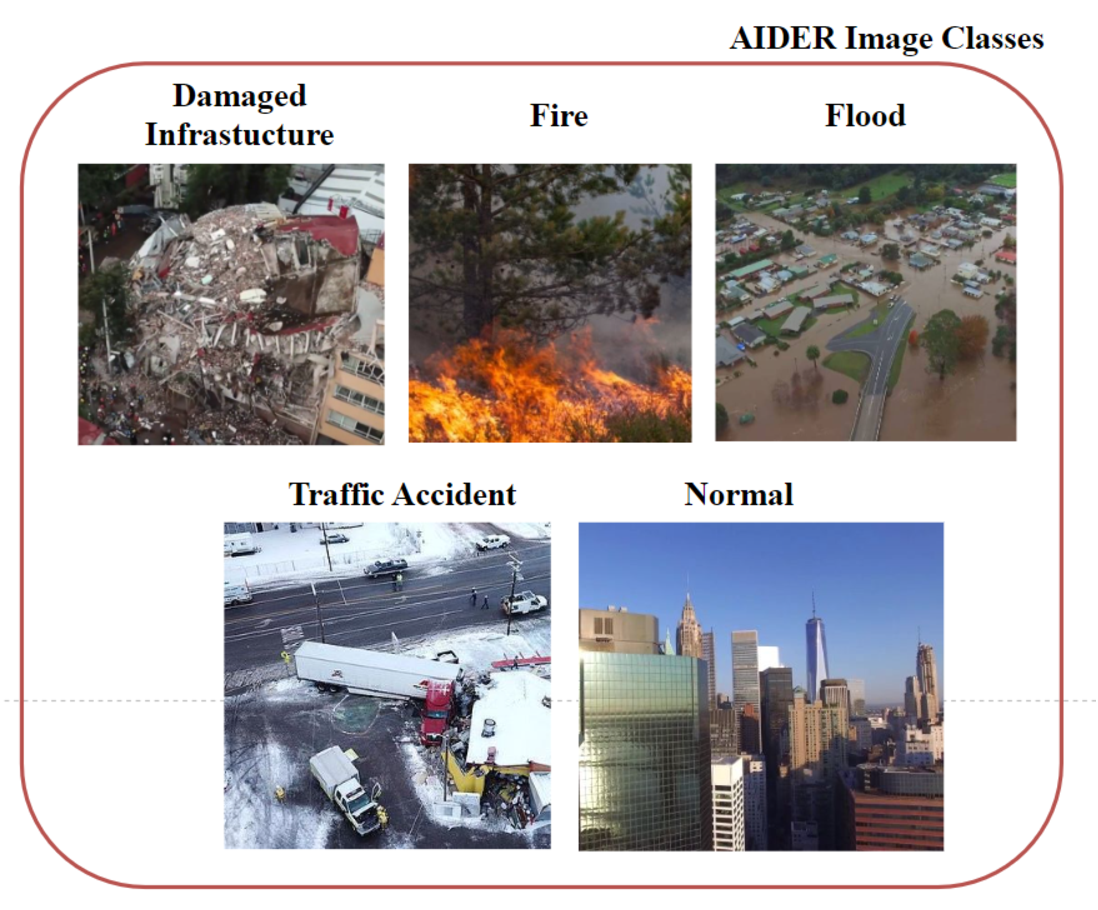}
    \caption{Example images from each AIDER subset disaster classes.}
    \label{fig:image_sample_AIDER}
\end{figure}
\begin{figure}[hbt!]
    \centering
    \includegraphics[height = 8.5cm, width=11.05cm]{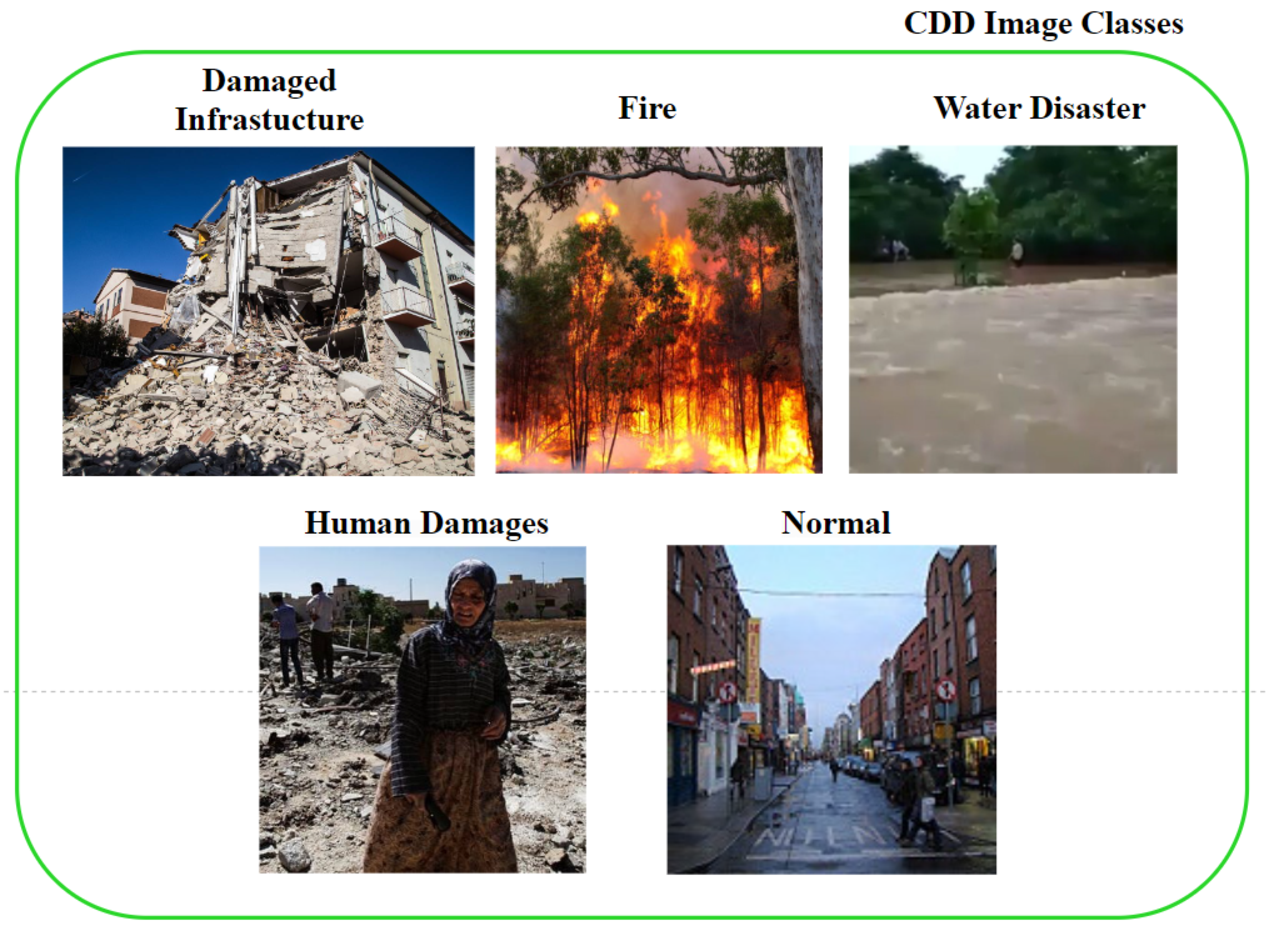}
    \caption{Example images from each CDD subset disaster classes.}
    \label{fig:image_sample_CDD}
\end{figure}
\begin{figure}[hbt!]
    \centering
    \includegraphics[height = 8.5cm, width=11.05cm]{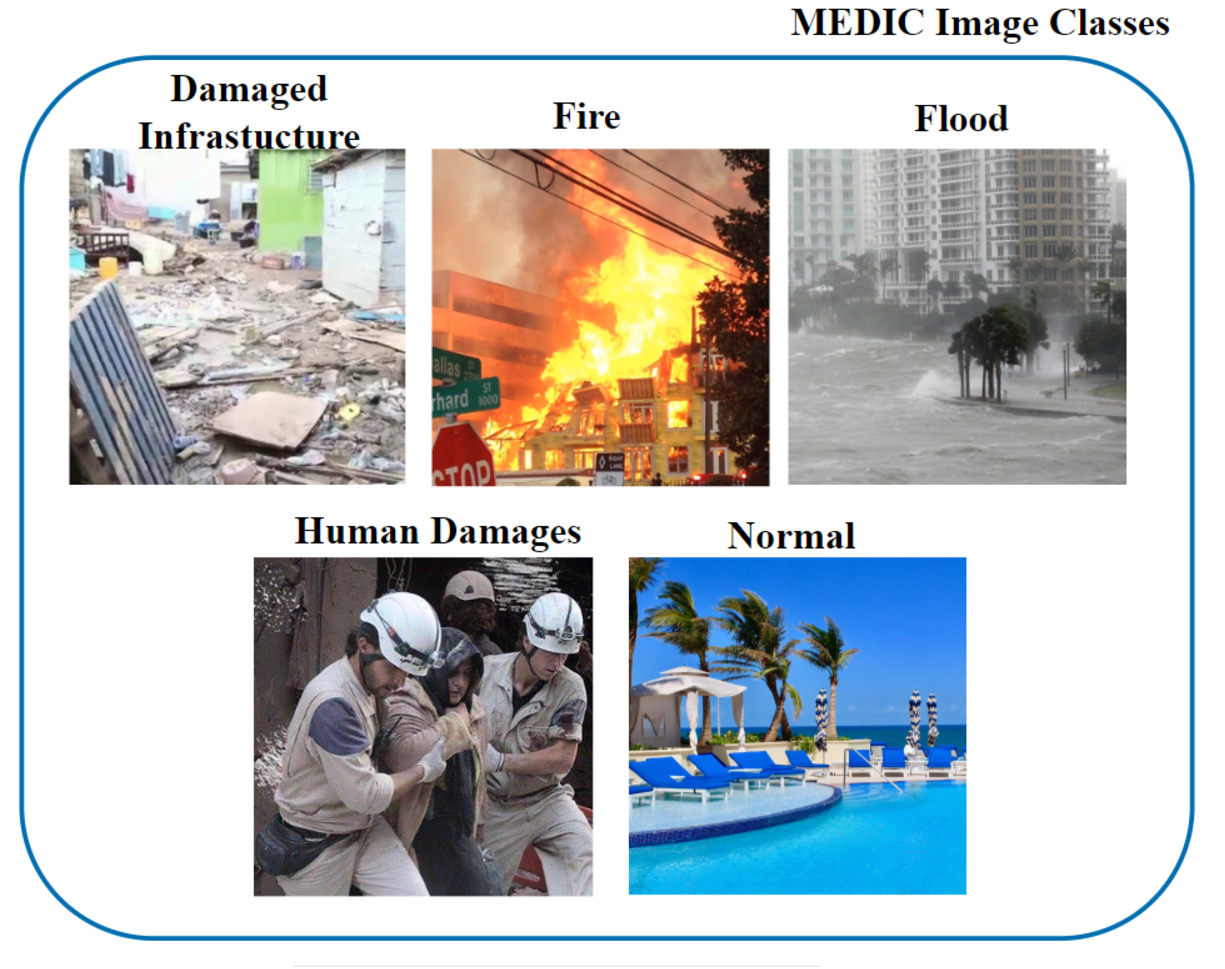}
    \caption{Example images from each MEDIC disaster classes.}
    \label{fig:image_sample_MEDIC}
\end{figure}

As mentioned in the introductory section, we utilized the AIDER, CDD, and the MEDIC dataset for the algorithmic baseline evaluation, but we also incorporated four FSL benchmarked (FC-100, CIFAR-FS, miniImageNet, tieredImageNet) to assess our model's performance in the benchmarked contexts to better analyze its robustness. Each dataset is divided into a train-valid-test ratio of 4:1:2. Except as otherwise indicated (see Table \ref{tab:hyperparams}), the ResNet12 architecture \cite{sun2019meta} is used as the CNN backbone throughout our experiment. Fig.\ref{fig:image_sample_AIDER}, \ref{fig:image_sample_CDD} and \ref{fig:image_sample_MEDIC} also depict representative samples of each disaster image class type for the respective aforementioned dataset. 

\subsection{Datasets}

For the benchmarked dataset we utilized a resolution of 32$\times$32$\times$3 for each image in the FC-100 and CIFAR-FS and a resolution of 84$\times$84$\times$3 for each image in the miniImageNet and tieredImageNet, in accord to their original dataset configuration. For the disaster datasets, we utilize a resolution of 128$\times$128$\times$3 for each image.

The choice of the image size for the disaster imagery appears to be non-optimal in the sense that essential spatial details that are key for identifying specific disaster cues may not be detected easily and hence leading to possible degradation of the critical classification task. We argued that our model is not designed to replace high-resolution disaster classification pipelines, but rather to provide a rapid few-shot scene-level classification when annotated data and computational resources are limited. Therefore, in contrast to fine-grained object-level details, we contend that the model is still capable of capturing the dominant spatial semantics and structural patterns that are most pertinent for scene-level classification tasks (e.g., large-scale layout, texture distribution, and contextual co-occurrence of objects) despite the resizing and cropping, and hence will not cause significant loss of the key spatial details. Such design decision is especially crucial for few-shot learning, since when there are few labeled examples available, operating at a lower resolution helps prevent overfitting, speeds up episodic training, and drastically lowers computing complexity.

As we will observe in the results section, we find that even under moderate downsampling to 128$\times$128, the model is still able to differentiate between disaster categories that are mainly defined by global spatial configurations (e.g., flooded regions, wildfire spread patterns, or urban damage layouts, espcially for the AIDER dataset). Moreover, similar resolution settings have been widely adopted in prior few-shot model (e.g., Meta-Album by Ullah et al. \cite{ullah2022meta}, which argues that such resolution offers a good compromise between computational burden and loss of accuracy), as well as their application in remote sensing (e.g., HDCPAA by Li et al. \cite{li2025hdcpaa}, few-shot classification via multi-scale relation distillation \cite{zhou2026few}) suggesting that such a trade-off between resolution and generalization is both practical and effective in low-data regimes.

\subsubsection*{\textbf{AIDER} \cite{kyrkou2020emergencynet}} This data set consisted of aerial images belonging to any of the four types of disasters, namely aftermath of collapsed buildings, fires, floods, and traffic accidents, as well as a non-disaster type (labeled as \emph{normal}). The latter class contained a significantly higher number of data samples than the disaster classes to replicate the real-world scenario as much as possible. The author only made a subset of the original dataset publicly available, and it contained 511 collapsed building images, 521 fire images, 526 flood images, 485 traffic images. and 4390 normal images, making a total of 6433 images, as illustrated in Table \ref{table:data_class_distribution_AIDER}. 

\begin{table}[hbt!]
    \centering
    \caption{The training, validation and test image sets for each AIDER subset class.}
    \vspace{0.3mm} 
\begin{tabular}{c|c|c|c|c}
 \hline
 \textbf{Class} & \textbf{Train} & \textbf{Valid} & \textbf{Test} & \textbf{Total per Class}\tabularnewline
 \hline
 Collapsed Building & 367 & 41 & 103 & 511\tabularnewline
 \hline
 Fire & 249 & 63 & 209 & 521\tabularnewline
  \hline
 Flood & 252 & 63 & 211 & 526\tabularnewline
  \hline
 Traffic & 232 & 59 & 194 & 485\tabularnewline
 \hline
Normal & 2107 & 527 & 1756 & 4390\tabularnewline
\hline
\textbf{Total Per Set} & \textbf{3207} & \textbf{753} & \textbf{2473} & \textbf{6433}\tabularnewline
 \hline
\end{tabular}
\label{table:data_class_distribution_AIDER}
\end{table}

\subsubsection*{\textbf{CDD} \cite{niloy2021novel}} This dataset contained not only the classes found in AIDER, but also new disaster classes such as land disasters (e.g., drought and landslides) and human-based disasters such as diseases and injuries. Also, unlike AIDER, some of the existing classes are also further sub-categorized. (For example, for the damaged infrastructure class, the images are distinguished on the basis of whether it is caused by an earthquake or some other factors). However, in our experiment, we considered all types of land disasters to belong to one class, all infrastructure damages to be in one class, and all human damages to be categorized into one class. Therefore, we considered CDD to be of 5 classes, similar to that of AIDER. This is to ensure a fairer comparison of each method in spite of the slight difference in the nature of dataset. There are a total of 1454 images belonging to the damaged infrastructure class, 933 images belonging to the fire class, 1035 images containing water-based disasters like flood, 657 human-based disasters (or damages), and 9237 non-disaster images, as illustrated in Table \ref{table:data_class_distribution_CDD}, making a total of 13316 images.

\begin{table}[hbt!]
    \centering
    \caption{The training, validation and test image sets for each CDD class.}
    \vspace{0.3mm} 
\begin{tabular}{c|c|c|c|c}
 \hline
 \textbf{Class} & \textbf{Train} & \textbf{Valid} & \textbf{Test} & \textbf{Total per Class}\tabularnewline
 \hline
 Infrastructure Damages & 1046 & 117 & 291 & 1454\tabularnewline
 \hline
 Fire & 447 & 112 & 374 & 933\tabularnewline
  \hline
 Water Disasters & 496 & 125 & 414 & 1035\tabularnewline
  \hline
 Human Damages & 315 & 79 & 263 & 657\tabularnewline
 \hline
Normal & 4433 & 1109 & 3695 & 9237\tabularnewline
\hline
\textbf{Total Per Set} & \textbf{6737} & \textbf{1542} & \textbf{5037} & \textbf{13316}\tabularnewline
 \hline
\end{tabular}
\label{table:data_class_distribution_CDD}
\end{table}

\subsubsection*{\textbf{MEDIC} \cite{alam2023medic}} This dataset is designed for humanitarian response and originally consisted of 71,198 images of a wide variety of disaster classes, some of which include earthquake, fire, flood, hurricane, landslide, as well as human damage and non-disaster classes. Unlike AIDER and CDD, MEDIC is designed for multi-task learning evaluation. This means that in addition to the disaster type images, information such as the type of damage resulted (e.g., infrastructure and/or utility damages), the severity of the damages, and whether a humanitarian response has been made. However, in our work we focused solely on a subset of the dataset that contains the disaster type as discussed in AIDER and CDD so fairer evaluations could be made for each FSL method relative to the other two datasets, which lack such additional information. Like CDD, we once again classed every human-based damage and disaster into one class, every earthquake and landslide incident under the class of infrastructure damages, and every flood and hurricane incident under the class of water disasters, thus considering MEDIC to be of five classes. Therefore, there are a total of 1418 infrastructure damage images, 349 fire disaster images, 396 water disaster images, 240 human disaster and damage images, and 2985 non-disaster images, as summarized in Table \ref{table:data_class_distribution_MEDIC}, making a total of 10152 images in our MEDIC subset.

\begin{table}[hbt!]
    \centering
    \caption{The training, validation and test image sets for each MEDIC subset class.}
    \vspace{0.3mm} 
\begin{tabular}{c|c|c|c|c}
 \hline
 \textbf{Class} & \textbf{Train} & \textbf{Valid} & \textbf{Test} & \textbf{Total per Class}\tabularnewline
 \hline
 Infrastructure Damages & 1020 & 114 & 284 & 1418 \tabularnewline
 \hline
 Fire & 167 & 42 & 140 & 349 \tabularnewline
  \hline
 Flood & 189 & 48 & 159 & 396 \tabularnewline
  \hline
 Human Damages & 115 & 29 & 96 & 240\tabularnewline
 \hline
Normal & 1432 & 359 & 1194 & 2985 \tabularnewline
\hline
\textbf{Total Per Set} & \textbf{2923} & \textbf{592} & \textbf{1873} & \textbf{10152}\tabularnewline
 \hline
\end{tabular}
\label{table:data_class_distribution_MEDIC}
\end{table}

\subsubsection*{\textbf{CIFAR-FS and FC-100}}

There are 60 meta-training classes, 20 meta-validation classes, and 20 meta-testing classes in the FC-100 dataset. Each class is comprised of 600 images. Few-shot learning is rendered difficult in this dataset as small image size are correlated with poor image quality. The CIFAR-FS dataset comprised of 20 meta-testing classes, 16 meta-validation classes, and 64 meta-training classes. Like FC-100, the few-shot tasks are rendered difficult again due to the significant intra-class similarity between the images.

\subsubsection*{\textbf{miniImageNet and tieredImageNet}}

The miniImageNet comprised of 600 images per class, which includes a broad range of images from the 100 classes of ImageNet. The meta-train-valid-test split consists of images of 64, 16, and 20 classes respectively. The tieredImageNet includes 351 classes for meta-training, 97 classes for meta-validation, and 160 classes for meta-testing. Like the miniImageNet, each class is comprised of 600 images. \\

\subsection{\textbf{Experimental Setup and Evaluations}}

To address the class imbalance presented in the aerial disaster datasets, we implemented under-sampling on the training and validation samples before training. All images in the respective dataset are re-scaled via division of each original pixel value by 255 as a pre-processing step. All experimental simulations for this work were performed using PyTorch library in Python utilizing Google’s Colab Pro+ Tesla T4 Graphical Processing Units (GPU). The number of training episodes utilized for our approach is 80000, and the optimizer utilized for our network is that of Adam with a learning rate of 0.001. Each training episode corresponds to an $N_c$-way $N_S$-shot task, where $N_c$ classes are uniformly sampled without replacement from the meta-training set. For each class, $N_S$ support samples and $N_Q$ query samples are randomly selected without overlap, forming the support and query sets respectively. In accord to the majority of FSL works, we chose $N_{c} = 5$, $N_{S} = 1$ or $5$, and hence $N_Q = 5$ or $25$. The random seed value is fixed to 42 for class sampling, data shuffling, and model initialization, and the batch composition is set to 16.

The loss functions utilized during the training of our selected models differ depending on the nature of the method. For the Siamese network, the loss function implemented is that of the contrastive loss ($\ell_{constrastive}$). For the other SOTAs, except the relation network, the loss function used is the categorical cross entropy ($\ell_{CCE}$), for which our $\ell_{ABHFA}$ has been incorporated as expressed in Eq. \eqref{eq6}. As mentioned in the Related Works section, the loss function implemented for the RelationNet is the MSE loss ($\ell_{MSE}$), which can be mathematically written as

\begin{equation} \label{eq8}
\begin{split}
\ell_{MSE} = argmin \sum_{i}^{m}\sum_{j}^{n} (r_{i,j} - \boldsymbol{1}(y_{i} ==y_{j}))^{2}.
\end{split}
\end{equation} 

The matched pairs $y_{i} ==y_{j}$ have a similarity score of 1 and the mismatched pairs have a similarity score of 0. $i$ and $j$ run from 1 to 5 (where $m$ and $n$ = 5).

\subsection{\textbf{Evaluation Metrics}}

For all datasets used, we used the Average Accuracy (AA) classification accuracy scores as the main metric to assess the classification effectiveness of each SOTAs and our approach. The accuracy for each class $c$ is mathematically defined as

\begin{equation} \label{eq9}
\begin{split}
Accuracy_{c} = \frac{TP_{c}+TN_{c}}{TP_{c}+FN_{c}+TN_{c}+FN_{c}},
\end{split}
\end{equation} 

and the AA is described as 

\begin{equation} \label{eq10}
\begin{split}
AA = \sum_{c=1}^{\mathcal{C}}Accuracy_{c} ,
\end{split}
\end{equation} 

where the variables $TP$, $FP$, $TN$, and $FN$ represent the true positive, false positive, true negative, and false negative classes respectively. $\mathcal{C}$ denotes the total number of classes. In addition, we also utilized the confusion matrix to analyze correct and incorrect classifications incurred by our model, as will be illustrated and elaborated in the Results section.

For the aerial disaster datasets, due to their significant class imbalanced nature, we have also evaluated our model via the precision, recall, and the macro F1-scores to provide a more comprehensive assessment. The F1-score for each class $c$ can be expressed in terms of the precision and recall metric (for each $c$) as

\begin{equation} \label{eq11}
    F1_{c} = \frac{2\cdot Precision_{c} \cdot Recall_{c}}{Precision_{c}+Recall_{c}}
\end{equation}

where

\begin{equation} \label{eq12}
    Precision_{c} = \frac{TP_{c}}{TP_{c}+FP_{c}},
\end{equation}

\begin{equation} \label{eq13}
    Recall_{c} = \frac{TP_{c}}{TP_{c}+FN_{c}}.
\end{equation}

Then the macro F1-score reads

\begin{equation} \label{eq14}
    F1_{macro} = \frac{1}{C}\sum_{c=1}^{\mathcal{C}}F1_{c}.
\end{equation}

For the classification efficiency comparison, we utilized the number of parameters and FLOPs for the benchmarked datasets, and additionally included inference time and memory usage for the AIDER, CDD and MEDIC evaluation.

\subsection{\textbf{Data Augmentations}}

In alignment with a majority of FSL works, we stressed the importance of data augmentation in our approach to attain SOTA few-shot classification performances. In light of this, for all the training set (benchmarked and aerial disaster datasets), we deployed the following image augmentations: center cropping (by 32/84/128$\times$32/84/128, depending on the datasets), resizing (by 32/84/128$\times$32/84/128, again depending on the datasets), random vertical and horizontal flips, random rotations by 90\degree, 180\degree and 270\degree, random horizontal flips followed by the same aformentioned random rotations, as well as random colour jitter, blurring and grayscaling. The probability of all augmentation that are performed in a random manner is set to 0.5.

\section{Results and Discussions}

All implemented algorithms underwent a comprehensive evaluation set of 10 runs/trials, and the results, encapsulated as the AA scores, as well as their corresponding 95$\%$ confidence interval, were duly recorded. For the disaster datasets, additional macro F1-scores and their confidence intervals were also reported. The obtained numerical scores, representing the performance of all the SOTA methodologies and our proposed approaches, were established as the foundational benchmarks for the trio of datasets under consideration. In order to provide a thorough and robust evaluation of each SOTA algorithm, two distinct evaluation methodologies were employed, as mentioned in the experimental setup section. They are denoted as the 5-way-1 shot and 5-way-5 shot evaluations, providing insights into the performance of the models under varying levels of data sparsity.

\subsection{Results on Benchmarked Datasets}

Table \ref{tab:CIFAR-FS}, \ref{tab:FC-100} and \ref{tab:miniImageNet_tierImageNet} depicts the selected SOTA scores with that of our approach on the CIFAR-FS, the FC-100, the miniImageNet and the tieredImageNet respectively, along with their corresponding required parameters and inference FLOPs (for 5-shot). The approaches selected for comparison were mostly the same as that by Roy et al. \cite{roy2022felmi}, except we incorporated the HELA-VFA \cite{lee2024hela}, ANROT-HELANet \cite{lee2025anrot}, and the Wasserstein few-shot variational inference model \cite{ambrogioni2018wasserstein}, works which involved variational few-shot classification via the Hellinger and Wasserstein distance respectively.

\begin{table}
\centering
\caption{The (average) classification \textbf{accuracy} (in $\%$), the corresponding 95$\%$ confidence interval for 10 trials, as well as the parameters usage and FLOPs (5-shot) for the 5-way 1-shot and 5-shot evaluations utilizing our ABHFA-Net relative to the SOTAs on the \textbf{CIFAR-FS}. The bolded values denote the best (highest) value obtained (except for the efficiency metrics).}
\begin{tabular}{p{3.5cm}p{2cm}p{2cm}p{1cm}p{2.5cm}}
\hline
\textbf{Methods} & \textbf{5-way-1-shot} & \textbf{5-way-5-shot} & \textbf{Params} & \textbf{FLOPs (5-shot)} \\
\hline
ProtoNet \cite{snell2017prototypical} & 72.7$\pm$0.7 & 83.5$\pm$0.5 & 12.4M & 104.6G \\
MetaOptNet \cite{lee2019meta} & 72.6$\pm$0.7 & 84.3$\pm$0.5 & 12.4M & 104.7G \\
DSN-MR \cite{tao2021graph} & 75.6$\pm$0.9 & 86.2$\pm$0.6 & 13.4M & 104.8G\\
RFS-Simple \cite{tian2020rethinking} & 71.5$\pm$0.8 & 86.0$\pm$0.5 & 12.5M & 104.6G \\
RFS-distill \cite{tian2020rethinking} & 73.9$\pm$0.8 & 86.9$\pm$0.5 & 12.5M & 104.6G\\
IER-distill \cite{rizve2021exploring} & 77.6$\pm$1.0 & 89.7$\pm$0.6 & 13.5M & 104.8G\\
PAL \cite{ma2021partner} & 77.1$\pm$0.7 & 88.0$\pm$0.5 & 12.5M & 209.2G \\
 SKD-Gen1 \cite{rajasegaran2020self} & 76.6$\pm$0.9 & 88.6$\pm$0.5 & 12.4M & 104.7G \\
 Label Halluc \cite{jian2022label} & 78.0$\pm$1.0 & 89.4$\pm$0.6 & 12.4M & 104.6G \\
 FeLMi \cite{roy2022felmi} & 78.2$\pm$0.7 & 89.5$\pm$0.5 & 12.4M & 104.6G\\
 HELA-VFA \cite{lee2024hela} & 78.9$\pm$0.4 & 90.7$\pm$0.7 & 13.6M & 104.8G\\
 ANROT-HELANet \cite{lee2025anrot} & 79.2$\pm$0.7 & 90.9$\pm$0.5 & 13.6M & 314.4G \\
 Wasserstein VI \cite{ambrogioni2018wasserstein} & 77.9$\pm$1.3 & 87.9$\pm$1.6 & 13.3M & 104.8G \\ 
 \textbf{ABHFA-Net} & \textbf{80.7$\pm$0.8} & \textbf{92.3$\pm$0.7} & 13.4M & 104.9G\\
 \hline
\end{tabular}
\label{tab:CIFAR-FS}
\begin{tablenotes}
    \item *All models are trained with the ResNet12 CNN backbone in a standardized environment.
\end{tablenotes}
\end{table}

\begin{table}
\centering
\caption{The (average) classification \textbf{accuracy} (in $\%$), the corresponding 95$\%$ confidence interval for 10 trials, as well as the parameters usage and FLOPs (5-shot) for the 5-way 1-shot and 5-shot evaluations utilizing our ABHFA-Net relative to the SOTAs on the \textbf{FC-100}. The bolded values denote the best (highest) value obtained (except for the efficiency metrics).}
\begin{tabular}{p{3.5cm}p{2cm}p{2cm}p{1cm}p{2.5cm}}
\hline
\textbf{Methods} & \textbf{5-way-1-shot} & \textbf{5-way-5-shot} & \textbf{Params} & \textbf{FLOPs (5-shot)} \\
\hline
 ProtoNet & 37.5$\pm$0.6 & 52.5$\pm$0.6 & 12.4M & 104.6G\\
 MetaOptNet & 41.1$\pm$0.6 & 55.5$\pm$0.6 & 12.4M & 104.7G \\
 TADAM \cite{oreshkin2018tadam} & 40.1$\pm$0.4 & 56.1$\pm$0.4 & 13.2M & 157.0G\\ 
 MTL \cite{sun2019meta} & 45.1$\pm$1.8 & 57.6$\pm$0.9 & 37.3M & 104.7G\\
 RFS-Simple & 42.6$\pm$0.7 & 59.1$\pm$0.6 & 12.5M & 104.6G\\
 Deep-EMD \cite{zhang2020deepemd} & 46.5$\pm$0.8 & 63.2$\pm$0.7 & 12.4M & 104.7G \\
 RFS-distill & 44.6$\pm$0.7 & 60.9$\pm$0.6 & 12.5M & 104.6G\\
 IER-distill & 48.1$\pm$0.8 & 65.0$\pm$0.7 & 13.5M & 104.8G\\
 PAL & 47.2$\pm$0.6 & 64.0$\pm$0.6 & 12.5M & 209.2G \\
 SKD-Gen1 & 46.5$\pm$0.8 & 64.2$\pm$0.8 & 12.4M & 104.7G \\
 AssoAlign \cite{afrasiyabi2020associative} & 45.8$\pm$0.5 & 59.7$\pm$0.6 & 11.2M & 111.0G \\
 InfoPatch \cite{gao2021contrastive} & 43.8$\pm$0.4 & 58.0$\pm$0.4 & 12.8M & 209.2G \\
 Label Halluc & 47.3$\pm$0.7 & 67.9$\pm$0.7 & 12.4M & 104.6G\\
 FeLMi & 49.0$\pm$0.7 & 68.7$\pm$0.7 & 12.4M & 104.6G\\
 HELA-VFA & 50.3$\pm$0.3 & 69.1$\pm$0.2 & 13.6M & 104.8G \\
 ANROT-HELANet & 51.2$\pm$0.6 & 69.6$\pm$0.6 & 13.6M & 314.4G\\ 
 Wasserstein VI & 50.6$\pm$1.9 & 68.5$\pm$1.5 & 13.3M & 104.8G \\ 
 \textbf{ABHFA-Net} & \textbf{51.7$\pm$0.4} & \textbf{70.5$\pm$0.6} & 13.4M & 104.9G \\
 \hline
\end{tabular}
\label{tab:FC-100}
\begin{tablenotes}
    \item *All models, except for AssoAlign which utilized the ResNet18, are trained with the ResNet12 CNN backbone in a standardized environment.
\end{tablenotes}
\end{table}

\begin{table*} 
\centering
\caption{The (average) classification \textbf{accuracy} (in $\%$), the corresponding 95$\%$ confidence interval for 10 trials, as well as the parameters usage and FLOPs (5-shot) for the 5-way 1-shot and 5-shot evaluations utilizing our ABHFA-Net relative to the SOTAs on the \textbf{miniImageNet} and \textbf{tieredImageNet}. The bolded values denote the best (highest) value obtained (except for the efficiency metrics).}
\begin{tabular}{p{3.0cm}p{1,5cm}p{1.5cm}p{1.5cm}p{1.5cm}p{1.0cm}p{1.5cm}}
\hline
\multicolumn{7}{|c|}{\textbf{Datasets}} \\ \hline
& \multicolumn{2}{|c|}{\textbf{miniImageNet}} & \multicolumn{2}{|c|}{\textbf{tieredImageNet}} &
\multicolumn{2}{|c|}{\textbf{Efficiency Metrics}}\\ 
\hline
\textbf{Methods} & \textbf{5-way-1-shot} & \textbf{5-way-5-shot} & \textbf{5-way-1-shot} & \textbf{5-way-5-shot} & \textbf{Params} & \textbf{FLOPs(5-shot)} \\
\hline
ProtoNet & 60.4$\pm$0.8 & 78.0$\pm$0.6 & 65.7$\pm$0.9 & 83.4$\pm$0.7 & 12.4M & 701.6G \\
MetaOptNet & 62.6$\pm$0.6 & 78.6$\pm$0.5 & 66.0$\pm$0.7 & 81.6$\pm$0.5 & 12.4M & 701.7G\\
MTL & 61.2$\pm$1.8 & 75.5$\pm$0.8 & 65.6$\pm$1.8 & 80.6$\pm$0.9 & 37.3M & 701.6G\\
TADAM & 58.5$\pm$0.3 & 76.7$\pm$0.3 & - & - & 13.2M & 1052.6G \\
Shot-Free \cite{ravichandran2019few} & 59.0$\pm$0.4 & 77.6$\pm$0.4 & 66.9$\pm$0.4 & 82.6$\pm$0.4 & 14.5M & 701.6G\\
Deep-EMD & 65.9$\pm$0.8 & 82.4$\pm$0.6 & 71.2$\pm$0.9 & 86.0$\pm$0.6 & 12.4M & 703.7G\\
FEAT \cite{ye2020few} & 66.8$\pm$0.2 & 82.1$\pm$0.1 & 70.8$\pm$0.2 & 84.8$\pm$0.2 & 16.1M & 701.6G\\
 DSN-MR & 64.6$\pm$0.7 & 79.5$\pm$0.5 & 67.4$\pm$0.8 & 82.9$\pm$0.6 & 13.4M & 706.8G\\
 Neg-Cosine \cite{liu2020negative} & 63.9$\pm$0.8 & 81.6$\pm$0.6 & - & - & 12.8M & 701.6G\\
 P-Transfer \cite{shen2021partial} & 64.2$\pm$0.8 & 80.4$\pm$0.6 & - & - & 12.4M & 701.6G\\
 MELR \cite{fei2021melr} & 67.4$\pm$0.4 & 83.4$\pm$0.3 & 72.1$\pm$0.5 & 87.0$\pm$0.4 & 16.1M & 702.1G \\
 TapNet \cite{yoon2019tapnet} & 61.7$\pm$0.2 & 76.4$\pm$0.1 & 63.1$\pm$0.2 & 80.3$\pm$0.1 & 13.2M & 701.7G \\
 IEPT \cite{zhang2021iept} & 67.1$\pm$0.4 & 82.9$\pm$0.3 & 72.2$\pm$0.5 & 86.7$\pm$0.3 & 12.8M & 701.7G\\
 RFS-Simple & 62.0$\pm$0.6 & 79.6$\pm$0.4 & 69.7$\pm$0.7 & 84.4$\pm$0.6 & 12.5M & 701.6G\\
 RFS-distill & 64.8$\pm$0.8 & 82.4$\pm$0.4 & 71.5$\pm$0.7 & 86.0$\pm$0.5 & 12.5M & 701.6G \\
 IER-distill & 66.9$\pm$0.8 & 84.5$\pm$0.5 & 72.7$\pm$0.9 & 86.6$\pm$0.8 & 13.5M & 701.8G  \\
 SKD-Gen1 & 66.5$\pm$1.0 & 83.2$\pm$0.5 & 72.4$\pm$1.2 & 86.0$\pm$0.6 & 12.4M & 701.7G\\
 AssoAlign & 60.0$\pm$0.7 & 80.4$\pm$0.7 & 69.3$\pm$0.6 & 86.0$\pm$0.5 & 11.2M & 783.5G\\
 Label Halluc & 67.0$\pm$0.7 & 85.9$\pm$0.5 & 72.0$\pm$0.9 & 86.8$\pm$0.6 & 12.4M & 701.6G \\
 FeLMi & 67.5$\pm$0.8 & 86.1$\pm$0.4 & 71.6$\pm$0.9 & 87.1$\pm$0.6 & 12.4M & 701.6G \\
 VFS & 57.2$\pm$0.3 & 61.2$\pm$0.3 & - & - & 13.2M & 701.8G \\
 HELA-VFA & 68.2$\pm$0.3 & 86.7$\pm$0.7 & 72.5$\pm$0.5 & 87.6$\pm$0.1 & 13.6M & 701.8G\\
 ANROT-HELANet & 69.4$\pm$0.3 & \textbf{88.1$\pm$0.4} & 75.3$\pm$0.2 & \textbf{89.5$\pm$0.8} & 13.6M & 2105.3G\\ 
 Wasserstein VI & 68.7$\pm$1.5 & 86.9$\pm$1.7 & 74.6$\pm$0.8 & 87.1$\pm$1.1 & 13.3M & 701.8G\\ 
 \textbf{ABHFA-Net} & \textbf{73.4$\pm$1.3} & 87.4$\pm$0.7 & \textbf{77.9$\pm$0.8} & 88.9$\pm$1.5 & 13.4M & 701.9G \\
 \hline
\end{tabular}
\label{tab:miniImageNet_tierImageNet}
 \begin{tablenotes}
    \item *All models, except for AssoAlign which utilized the ResNet18, are trained with the ResNet12 CNN backbone in a standardized environment.
    \end{tablenotes}
\end{table*}

Firstly, we can observe that the AA obtained using the 5-way-5-shot approach is generally higher than that obtained using the 5-way-1-shot approach for all the methods utilized. This is because of the increased difficulty that can arise in utilizing FSL with fewer shots, since there exists less supporting information in the support set to guide the model in making accurate predictions. Secondly, for both 5-way-1-shot and 5-way-5-shot, all methods yielded lower classification accuracies for FC-100 relative to CIFAR-FS. This implies that the FC-100 indeed presented a larger obstacle to accurate classification for FSL than CIFAR-FS. Based on the accuracy statistics for miniImageNet and tieredImageNet, the two datasets exhibited varying degrees of difficulty for the few-shot classification, with the tieredImageNet posting a lesser challenge than miniImageNet as the former's AA values are generally higher. This is due to the greater amount of meta-training or base classes (351) in the tieredImageNet (as compared to 64 base classes in miniImageNet) that facilitated richer feature space representation, as well as the lesser semantic overlaps between the base and novel classes and thus cleaner task separation. These factors enabled better embedding and consequently leading to more stable prototypes and enhanced classification in the dataset. Lastly, our ABHFA-Net outperformed the SOTAs on almost all the benchmarked datasets, even for models utilizing Hellinger distance (HELA-VFA, ANROT-HELANet) and Wasserstein distance (Wasserstein VI), albeit with slightly higher improvement. For example, our model outperforms ANROT-HELANet (second-best model) in the 5-way-1-shot evaluation by 1.5$\%$, and in the 5-way-5-shot evaluation by 1.4$\%$ in the CIFAR-FS evaluation. Concurrently, our model outperforms the ANROT-HELANet again in the 5-way-1-shot and 5-way-5-shot evaluation on the FC-100 by 0.5$\%$ and 0.9$\%$ respectively. Finally, for the miniImageNet and tieredImageNet, our ABHFA-Net once again outperformed all selected approaches in Table \ref{tab:miniImageNet_tierImageNet} in the 5-way-1-shot evaluation but not on the 5-way-5-shot. Although our method outperformed the ANROT-HELANet in these dataset by 4.0$\%$ and 2.6$\%$ in the 1-shot comparison of miniImageNet and tieredImageNet respectively, it falls behind slightly by 0.7$\%$ and 0.6$\%$ relative to the same model in their 5-shot comparison.

The consistent superior performance of ABHFA-Net in both shot setting in general can be attributed to the improved estimation of latent Gaussian class distributions. Since the proposed Bhattacharyya-distance-based variational framework jointly accounts for both mean separation and covariance overlap (unlike the Waserstein, KL and Hellinger distance-based approaches), the model is able to better capture intra-class variability, uncertainty structure, and correlated latent features under richer support statistics. This enables distinctive probabilistic prototypes to be learned and consequently enhanced class separability and the classification scores. However, in miniImageNet, less stable covariance estimation and higher uncertainty in the latent representation space resulted due to the higher class semantic overlaps in the dataset. Additional anomalous or noisy support samples may disproportionately affect the prototype estimation in very low-shot context, as the ABHFA-Net depends heavily upon the distribution overlaps and covariance structure. Therefore, methods less sensitive to the covariance structure, such as the ANROT-HELANet may exhibit slightly stronger robustness in such context, explaining the trend in Table \ref{tab:miniImageNet_tierImageNet}.

On the efficiency aspect, we observed that most metric-learning and prototype-based approaches (ProtoNet, MetaOptNet, RFS-Simple, IER-distill, Label Hallucination, FelMi, HELA-VFA, Wasserstein VI, and ABHFA-Net) require only 12–14 million parameters and approximately 100–105/702 GFLOPs on CIFAR-FS/FC-100 and miniImageNet/tieredImageNet respectively, indicating that these methods share a similar ResNet-12 backbone and hence computational budget across all benchmarks. In contrast, methods such as TADAM, MTL, PAL and ANROT-HELANet has substantially higher computational costs due to additional adaptation, routing, or attention modules. The results in all the benchmarks demonstrate that increasing model capacity alone does not necessarily improve few-shot generalization. Specifically,  ABHFA-Net's consistent superior accuracy is not achieved through a larger backbone or substantially higher computational complexity, but rather through more effective feature aggregation and distance learning mechanisms. Unlike methods that improve performance through substantial increases in network depth, adaptation modules, or attention mechanisms, ABHFA-Net achieves SOTA accuracy under nearly the same computational budget as Wasserstein VI and HELA-VFA, models which strikes a favorable balance between predictive performance and computational efficiency. This also makes ABHFA-Net particularly attractive for resource-constrained deployment scenarios such as UAV-based remote sensing and edge-assisted disaster monitoring.

\subsection{Results on Aerial Disaster Image Datasets}

Unlike the benchmarked datasets, we do not have sufficient references for extracting the reported FSL classification values for the AIDER, CDD and MEDIC datasets \emph{a priori}. In view of this, we needed to perform training and inference for each of the compared models on each of the three datasets. The relevant training hyperparameters such as the backbone and optimizer used, Learning Rate (LR), and training episodes/epochs are listed in Table \ref{tab:hyperparams} for the displayed models, and were based on that of the corresponding prior literature as closed as possible.

Unlike the SOTA evaluation so far, it is important to note a distinction for the Siamese and triplet network. Given the unique structure of this network, which compares only a single pair and two pairs of images per episode of training respectively, results are exclusively reported for the 5-way-1 shot evaluation. This restriction aligns with the inherent characteristics of the Siamese and triplet network, elucidating its capabilities within the stipulated parameters of the experimental setup.

\begin{table}[t]
\centering
\caption{Training details (backbone and hyperparameters) for all evaluated methods on AIDER, CDD, and MEDIC datasets. All methods are trained under a standardized experimental setup unless otherwise specified.}
\label{tab:hyperparams}
\begin{tabular}{lccc}
\hline
\textbf{Method} & \textbf{Backbone} & \textbf{Optimizer and LR} & \textbf{Training Episodes}\\
\hline
Siamese Net \cite{koch2015siamese} & Conv-4 & SGD($5\times10^{-1}$) & 20000 \\
Triplet Net \cite{hoffer2015deep} & Conv-4 & SGD($5\times10^{-1}$) & 20000\\
ProtoNet & ResNet12 & Adam ($10^{-3}$) & 20000\\
RelationNet \cite{sung2018learning} & U-Net & SGD($10^{-3}$) & 10000\\
MatchingNet \cite{vinyals2016matching} & ResNet12 & SGD($10^{-3}$) & 10000 \\
SimpleShot \cite{wang2019simpleshot} & ResNet12 & SGD ($10^{-1}$) & 10000 \\
BDCSPN \cite{liu2020prototype} & WRN-28-10 & SGD ($10^{-1}$) & 600  \\
TIM \cite{boudiaf2008transductive} & WRN28-10 & Adam ($10^{-3}$) & 10,000 \\
Proto FT \cite{chen2019closer} & Conv-4 & Adam ($10^{-3}$) & 40000 \\
Transductive FT \cite{dhillon2019baseline} & WRN-28-10 & Adam ($10^{-3}$) & 40000 \\
VFA-Net \cite{han2023few} & ResNet12 & SGD($10^{-2}$) & 20000 \\ 
Wasserstein VI & ResNet12 & Adam ($10^{-3}$) & 80000 \\
HELA-VFA & ResNet12 & Adam ($10^{-3}$) & 80000 \\
CVAE \cite{xu2022generating} & ResNet12 & Adam ($10^{-4}$) & 2000 \\
FSDM \cite{giannone2022few} & DDPM & Adam ($2\times10^{-4}$) & 100000 \\
GL-ViT \cite{sun2022global} & ViT-S/16 & AdamW ($10^{-4}$) & 10000 \\
HC-Transformer \cite{he2022attribute} & ViT-S/8 & AdamW ($10^{-3}$) & 10000 \\
SSL-ViT-16 \cite{bhattacharyya2022visual} & ViT/16 & AdamW ($10^{-3}$) & 10000 \\
\textbf{ABHFA-Net} & ResNet12 & Adam ($10^{-3}$) & 80000 \\
\hline
\end{tabular}
\end{table}

Tables \ref{tab:final_results_AIDER}, \ref{tab:final_results_CDD} and \ref{tab:final_results_MEDIC} illustrate the selected SOTA scores with that of our approach for the AIDER, CDD and MEDIC datasets respectively, with their corresponding required parameters, peak GPU memory usage, inference FLOPs and time depicted in Table \ref{tab:final_results_efficiency} (all for 5-shot, except for Siamese and triplet network in which 1-shot values were used). For the precision and recall values for each approach, we dedicated them to Appendix D. From the accuracy values displayed across the three tables, we once again observe that both the F1-scores and the AA obtained using the 5-way-5-shot approach are generally higher than the 5-way-1-shot approach for all the methods utilized. This illustrates that the trend observed from the benchmarked is also generally true for a domain-specific context, along with the same reasons behind such observation. Nevertheless, our ABHFA-Net once again outperformed the evaluated approaches in all 3 datasets for the 5-way-1-shot (once again surpassing models utilizing the Hellinger distance (HELA-VFA) and Wasserstein distance (Wasserstein VI)). For the 5-way-5-shot, the HC-Transformer outperformed ours on the AIDER and CDD respectively, while the FSDM outperformed ours on the MEDIC. HC-Transformer has stronger feature representation and more explicit attribute-level learning, which better disentangled features across classes to improve their separability, and as a transformer-based approach, also benefits from higher-data regime due to their capability to model global attention better than our approach, which utilized mainly CNN-based encoder. This also explains why transformer-based methods like GL-ViT, HC-Transformer and SSL-ViT-16 in general yielded higher AA and F1-scores for all the three datasets relative to the rest, since the advantages of stronger feature learning conferred by transformer-based models in the presence of more support samples can slightly outweigh the benefits of distribution-based modeling in the higher-shot setting. Still, in the lower-shot context, our method yielded higher classification scores in a consistent manner.

\begin{table}
\centering
\caption{The \textbf{average accuracy (AA)}, the macro-F1 score, as well as the corresponding 95$\%$ confidence interval for 10 trials using the 5-way 1-shot and 5-way 5-shot learning evaluation on our AIDER subset simulation. The bolded values denote the best (highest) value obtained. \label{tab7}}
\begin{tabular}{p{3cm}p{2cm}p{2cm}p{2cm}p{2cm}}
\hline
\textbf{Methods} & \textbf{5-way-1-shot (AA)} & \textbf{5-way-1-shot (F1)} & \textbf{5-way-5-shot (AA)} & \textbf{5-way-5-shot (F1)} \\
\hline
 Siamese Network & 62.2$\pm$0.16 & 65.3$\pm$0.12 & - & - \\
 Triplet Network & 65.3$\pm$0.77 & 68.3$\pm$0.72 & - & -\\
 ProtoNet & 55.6$\pm$0.62 & 58.7$\pm$0.61 & 73.8$\pm$0.25 & 77.8$\pm$0.23\\
 RelationNet  & 44.4$\pm$0.64 & 48.3$\pm$0.59 & 75.9$\pm$0.16 & 79.4$\pm$0.14 \\
 MatchingNet & 36.2$\pm$0.21 & 41.5$\pm$0.26 & 73.2$\pm$1.10 & 76.1$\pm$1.05 \\
 SimpleShot  & 55.6$\pm$0.62 & 59.4$\pm$0.57 & 73.8$\pm$0.25 & 76.5$\pm$0.19 \\
 BDCSPN & 45.4$\pm$0.65 & 49.9$\pm$0.67 & 61.8$\pm$0.20 & 65.6$\pm$0.23\\
 TIM & 59.7$\pm$1.82 & 62.9$\pm$1.73 & 67.2$\pm$1.23 & 71.2$\pm$1.14 \\
 Prototype FT & 57.5$\pm$0.52 & 61.2$\pm$0.44 & 64.3$\pm$0.82 & 68.7$\pm$0.75 \\
 Transductive FT & 56.7$\pm$1.23 & 60.8$\pm$1.05 & 63.5$\pm$1.26 & 67.4$\pm$1.12 \\
 VFA-Net & 64.2$\pm$0.56 & 67.5$\pm$0.44 & 71.2$\pm$0.45 & 75.5$\pm$0.47\\
 Wasserstein VI & 62.7$\pm$0.99 & 64.4$\pm$0.96 & 66.1$\pm$1.16 & 69.6$\pm$1.04 \\
 HELA-VFA & 67.7$\pm$0.78 & 70.5$\pm$0.62 & 75.7$\pm$0.98 & 78.8$\pm$0.91 \\
 CVAE & 66.2$\pm$0.51 & 70.2$\pm$0.56 & 73.1$\pm$0.82 & 77.6$\pm$0.88 \\
 FSDM & 68.0$\pm$0.43 & 71.9$\pm$0.40 & 78.5$\pm$0.53 & 81.8$\pm$0.52\\
 GL-ViT & 66.7$\pm$0.70 & 70.8$\pm$0.75 & 76.8$\pm$0.64 & 80.1$\pm$0.52\\
 HC-Transformer & 67.8$\pm$0.57 & 71.8$\pm$0.45 & \textbf{79.4$\pm$0.87} & \textbf{82.5$\pm$0.81} \\
 SSL-ViT-16 & 66.9$\pm$0.84 & 71.1$\pm$0.72 & 77.2$\pm$0.65 & 80.7$\pm$0.74  \\
 \textbf{ABHFA-Net} & \textbf{68.2$\pm$1.19} & \textbf{72.2$\pm$1.05} & 78.3$\pm$0.75 & 81.4$\pm$0.67\\
 \hline
\end{tabular}
\label{tab:final_results_AIDER}
 \begin{tablenotes}
    \item *The backbone used for each of the method was listed in Table \ref{tab:hyperparams}.
    \end{tablenotes}
\end{table}

\begin{table}
\centering
\caption{The \textbf{average accuracy (AA)}, the macro-F1 score, as well as the corresponding 95$\%$ confidence interval for 10 trials using the 5-way 1-shot and 5-way 5-shot learning evaluation on the CDD dataset. The bolded values denote the best (highest) value obtained. \label{tab8}}
\begin{tabular}{p{3cm}p{2cm}p{2cm}p{2cm}p{2cm}}
\hline
\textbf{Methods} & \textbf{5-way-1-shot (AA)} & \textbf{5-way-1-shot (F1)} & \textbf{5-way-5-shot (AA)} & \textbf{5-way-5-shot (F1)} \\
\hline
 Siamese Network & 61.7$\pm$0.56 & 63.2$\pm$0.61 & - & -\\
 Triplet Network  & 62.2$\pm$0.63 & 64.5$\pm$0.55 & - & - \\
 ProtoNet & 55.5$\pm$1.19 & 57.7$\pm$1.12 & 69.5$\pm$0.98 & 70.4$\pm$1.07 \\
 RelationNet & 34.9$\pm$0.37 & 38.7$\pm$0.35 & 73.8$\pm$0.62 & 75.4$\pm$0.55 \\
 MatchingNet & 60.8$\pm$0.74 & 62.1$\pm$0.84 & 70.0$\pm$0.65 & 71.9$\pm$0.57 \\
 SimpleShot & 42.7$\pm$1.48 & 43.9$\pm$1.40 & 54.5$\pm$0.52 & 56.7$\pm$0.64 \\
 BDCSPN & 40.6$\pm$0.98 & 42.1$\pm$1.11 & 53.6$\pm$0.56 & 55.9$\pm$0.48 \\
 TIM & 57.8$\pm$0.36 & 59.2$\pm$0.51 & 66.7$\pm$0.71 & 67.4$\pm$0.89 \\
 Prototype FT & 55.4$\pm$0.74 & 58.3$\pm$0.89 & 65.5$\pm$0.26 & 68.0$\pm$0.45 \\
 Transductive FT & 52.7$\pm$0.42 & 55.1$\pm$0.56 & 64.2$\pm$0.30 & 67.9$\pm$0.41 \\
 VFA-Net & 62.8$\pm$0.17 & 64.2$\pm$0.32 & 69.4$\pm$0.59 & 72.2$\pm$0.46 \\
 Wasserstein VI & 61.2$\pm$1.18 & 64.5$\pm$1.05 & 65.8$\pm$1.08 &  68.4$\pm$1.21 \\
 HELA-VFA & 64.3$\pm$0.73 & 66.8$\pm$0.42 & 73.0$\pm$0.51 & 75.8$\pm$0.54\\
 CVAE & 63.1$\pm$0.89 & 65.5$\pm$0.82 & 71.3$\pm$0.75 & 73.6$\pm$0.81 \\
 FSDM & 64.4$\pm$0.56 & 66.5$\pm$0.62 & 75.4$\pm$0.88 & 77.2$\pm$0.92 \\
 GL-ViT & 61.5$\pm$0.78 & 64.6$\pm$0.93 & 73.2$\pm$0.81 & 75.5$\pm$0.86 \\
 HC-Transformer & 64.8$\pm$0.90 & 66.9$\pm$0.88 & \textbf{77.7$\pm$1.12} & \textbf{79.2$\pm$1.26} \\
 SSL-ViT-16 & 63.8$\pm$1.07 & 65.5$\pm$0.99 & 73.6$\pm$1.17 & 76.1$\pm$1.12\\
 \textbf{ABHFA-Net} & \textbf{65.0$\pm$0.95} & \textbf{67.1$\pm$1.14} & 74.2$\pm$0.65 & 77.6$\pm$0.73 \\
 \hline 
\end{tabular}
\label{tab:final_results_CDD}
 \begin{tablenotes}
    \item *The backbone used for each of the method was listed in Table \ref{tab:hyperparams}.
    \end{tablenotes}
\end{table}

\begin{table}
\caption{The \textbf{average accuracy (AA)}, the macro-F1 score, as well as the corresponding 95$\%$ confidence interval for 10 trials using the 5-way 1-shot and 5-way 5-shot learning evaluation on the MEDIC subset dataset. The bolded values denote the best (highest) value obtained. \label{tab9}}
\begin{tabular}{p{3cm}p{2cm}p{2cm}p{2cm}p{2cm}}
\hline
\textbf{Methods} & \textbf{5-way-1-shot (AA)} & \textbf{5-way-1-shot (F1)} & \textbf{5-way-5-shot (AA)} & \textbf{5-way-5-shot (F1)} \\
\hline
 Siamese Network & 57.3$\pm$1.62 & 60.5$\pm$1.57 & - & -\\
 Triplet Network & 55.3$\pm$1.72 & 58.4$\pm$1.56 & - & -\\
 ProtoNet & 34.3$\pm$1.29 & 37.1$\pm$1.38 & 55.9$\pm$1.66 & 57.9$\pm$1.57 \\
 RelationNet & 26.3$\pm$1.13 & 28.7$\pm$1.31 & 51.3$\pm$1.27 & 53.3$\pm$1.40\\
 MatchingNet & 38.6$\pm$1.01 & 40.5$\pm$1.03 & 47.0$\pm$1.36 & 48.9$\pm$1.21\\
 SimpleShot & 44.0$\pm$0.77 & 46.3$\pm$0.84 & 56.2$\pm$0.43 & 57.8$\pm$0.56 \\
 BDCSPN  & 40.7$\pm$0.54 & 43.6$\pm$0.80 & 59.6$\pm$0.24 & 61.7$\pm$0.57 \\
 TIM & 52.8$\pm$0.65 & 54.5$\pm$0.71 & 61.1$\pm$1.47 & 63.7$\pm$1.25 \\
 Prototype FT & 42.0$\pm$1.67 & 44.2$\pm$1.88 & 60.7$\pm$0.53 & 62.3$\pm$0.61 \\
 Transductive FT & 44.9$\pm$0.44 & 46.4$\pm$0.60 & 62.0$\pm$1.08 & 63.9$\pm$1.20 \\
 VFA-Net & 53.6$\pm$0.92 & 55.0$\pm$1.00 & 62.3$\pm$1.25 & 63.5$\pm$1.38 \\
 Wasserstein VI & 55.9$\pm$1.09 & 57.0$\pm$1.02 & 59.8$\pm$1.68 & 62.9$\pm$1.44 \\
 HELA-VFA & 58.4$\pm$0.96 & 60.9$\pm$1.21 & 66.1$\pm$1.56 & 68.2$\pm$1.83 \\
 CVAE & 58.7$\pm$1.32 & 60.6$\pm$1.22 & 65.8$\pm$1.48 & 66.4$\pm$1.39 \\
 FSDM & 59.9$\pm$1.25 & 60.5$\pm$1.18 & \textbf{67.3$\pm$1.15} & \textbf{70.5$\pm$1.22} \\
 GL-ViT & 58.3$\pm$1.37 & 60.3$\pm$1.30 & 63.5$\pm$1.17 & 65.6$\pm$1.05 \\
 HC-Transformer & 59.5$\pm$0.90 & 62.2$\pm$0.98 & 66.3$\pm$0.81 & 68.5$\pm$0.75 \\
 SSL-ViT-16 & 59.7$\pm$1.14 & \textbf{61.9$\pm$1.09} & 65.4$\pm$1.37 & 66.4$\pm$1.34 \\
 \textbf{ABHFA-Net} & \textbf{60.2$\pm$1.22} & \textbf{61.9$\pm$1.15} & 66.5$\pm$0.95 & 68.7$\pm$1.11 \\
 \hline
\end{tabular}
\label{tab:final_results_MEDIC}
 \begin{tablenotes}
    \item *The backbone used for each of the method was listed in Table \ref{tab:hyperparams}.
    \end{tablenotes}
\end{table}


\begin{table}
\caption{The efficiency metrics (Parameters, FLOPs (5-shot except for Siamese and Triplet Network), inference time (5-shot, per episode/per image, except for Siamese and Triplet Network), and peak memory usage (5-shot, except for Siamese and Triplet Network)) for each of the selected models in Table \ref{tab:final_results_AIDER}, \ref{tab:final_results_CDD}, and \ref{tab:final_results_MEDIC}. The bolded values denote the best (in this context, lowest FLOPs, parameters, inference time and memory requirement) values obtained.}
\begin{tabular}{p{3.7cm}p{1cm}p{1cm}p{3cm}p{2.0cm}}
\hline
\textbf{Methods} & \textbf{FLOPs} & \textbf{Params} & \textbf{Inference Time (per episode/per image)} & \textbf{Memory} \\
\hline
 Siamese Network (1-shot) & 13.9G & 2.5M & \textbf{0.17ms/0.04ms} & \textbf{177.3MB} \\
 Triplet Network (1-shot) & \textbf{13.8G} & \textbf{2.2M} & 5.04ms/0.17ms & 183.7MB \\
 ProtoNet & 838.1G & 12.8M & 83.78ms/1.68ms & 683.1MB \\
 RelationNet & 112.3G & 3.9M & 16.70ms/0.26ms & 261.0MB\\
 MatchingNet & 838.1G & 12.8M & 83.83ms/1.68ms & 656.3MB\\
 SimpleShot & 838.1G & 12.8M & 83.83ms/1.68ms & 651.6MB \\
 BDCSPN & 8389.4G & 36.9M & 3613.51ms/72.27ms & 970.2MB \\
 TIM & 8394.0G & 36.9M & 645.71ms/12.91ms & 13958.8MB \\
 Prototype FT & 369.1G & \textbf{2.2M} & 2000.11ms/40.00ms & 1620.2MB \\
 Transductive FT & 58757.0G & 36.9M & 1814.08ms/36.28ms & 1438.5MB \\
 VFA-Net & 838.2G & 15.7M & 84.14ms/1.68ms & 658.3MB \\
 Wasserstein VI & 838.1G & 13.2M & 83.91ms/1.68ms & 646.5MB \\
 HELA-VFA & 838.2G & 13.3M & 83.98ms/1.68ms & 646.5MB \\
 CVAE & 838.2G & 13.9M & 84.00ms/1.68ms & 797.7MB \\
 FSDM & 26630.2G & 8.6M & 3232.39ms/129.30ms & 802.1MB \\
 GL-ViT & 140.7G & 21.9M & 60.18ms/1.20ms & 298.9MB \\
 HC-Transformer & 414.7G & 21.9M & 146.13ms/2.92ms &  535.1MB \\
 SSL-ViT-16 & 414.7G & 21.9M & 356.22ms/2.37ms & 471.9MB \\
 \textbf{ABHFA-Net} & 838.8G & 13.3M & 94.02ms/1.88ms & 897.0MB \\
 \hline
\end{tabular}
\label{tab:final_results_efficiency}
 \begin{tablenotes}
    \item *The backbone used for each of the method was listed in Table \ref{tab:hyperparams}.
    \end{tablenotes}
\end{table}

We also noticed that the F1 scores are generally higher than the AA values for all datasets and for all methods compared. While accuracy is dominated by the dominant class (normal) in the datasets, macro-F1 assigns equal importance to each class. Consequently, moderate performance across minority disaster classes can yield higher macro-F1 values even when overall 
accuracy is lower. However, the gaps between the AA and macro-f1 scores in CDD (Table \ref{tab:final_results_CDD}) and MEDIC (Table \ref{tab:final_results_MEDIC}) are smaller than that of AIDER (Table \ref{tab:final_results_AIDER}). While differences of $\sim$ 4-5$\%$ between the AA and F1 scores for both shot setting in AIDER were reported, that difference was reduced to about $\sim$ 2-3$\%$ for both shot setting in CDD in general, and further reduced to about $\sim$ 1-2$\%$ for both shot setting in MEDIC in general.

Again, on the efficiency aspect, our ABHFA-Net once again maintains a favorable balance between classification performance and computational efficiency on all of the three datasets. Our model in these cases requires only 13.3M parameters and 838.8G FLOPs, which are comparable to conventional metric-based methods such as ProtoNet, MatchingNet, HELA-VFA, and Wasserstein VI. Although the incorporation of attention-guided feature aggregation increases inference time from approximately 84 ms to 95 ms per image and raises memory usage to 897 MB, these overheads remain moderate relative to the substantial gains reported in Tables \ref{tab:CIFAR-FS} - \ref{tab:miniImageNet_tierImageNet}. Furthermore, compared with transductive approaches such as TIM, BDCSPN, and Transductive Fine-Tuning, ABHFA-Net achieves significantly lower inference latency while avoiding expensive test-time optimization procedures. These results once again justify our stand that ABHFA-Net offers an attractive accuracy–efficiency trade-off, rendering it suitable for near-real-time UAV-based disaster monitoring scenarios.

The observed reduction in the gap between average accuracy (AA) and macro-F1 from AIDER to CDD and MEDIC can be attributed to differences in dataset characteristics. In AIDER, the relatively clearer class boundaries allow the model to achieve uneven performance across classes, where strong performance 
on minority classes combined with moderate errors on the dominant class leads 
to higher macro-F1 relative to accuracy. In contrast, CDD and especially MEDIC exhibit increased intra-class variability  and stronger semantic overlap between classes, which results in more uniformly distributed classification errors across all categories. Consequently, both accuracy and macro-F1 decrease in a similar manner, leading to a smaller gap between the two metrics. This trend indicates that as class ambiguity increases, macro-F1 becomes less  inflated relative to accuracy, reflecting more balanced but overall more challenging classification performance.


\begin{figure}[hbt!]
    \centering   
    \includegraphics[scale=0.76]{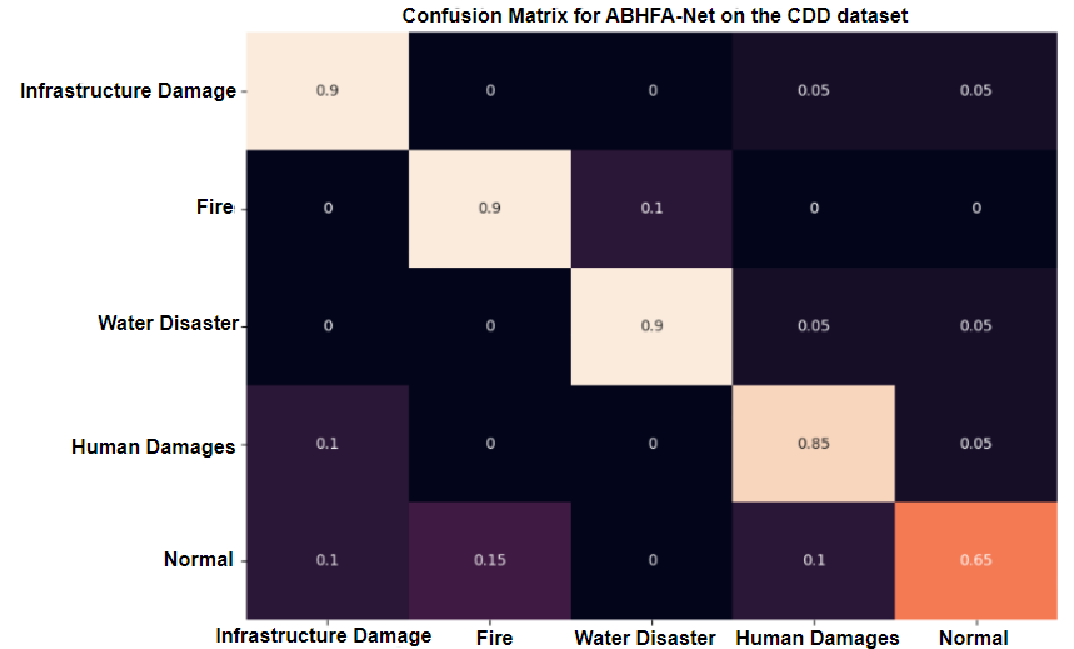}
    \caption{The confusion matrix generated for the 5-way-5-shot approach  for our ABHFA-Net on the CDD.}
    \label{fig:confused_matrix_CDD2}
\end{figure}


\begin{figure}[hbt!]
    \centering
    \includegraphics[scale=0.76]{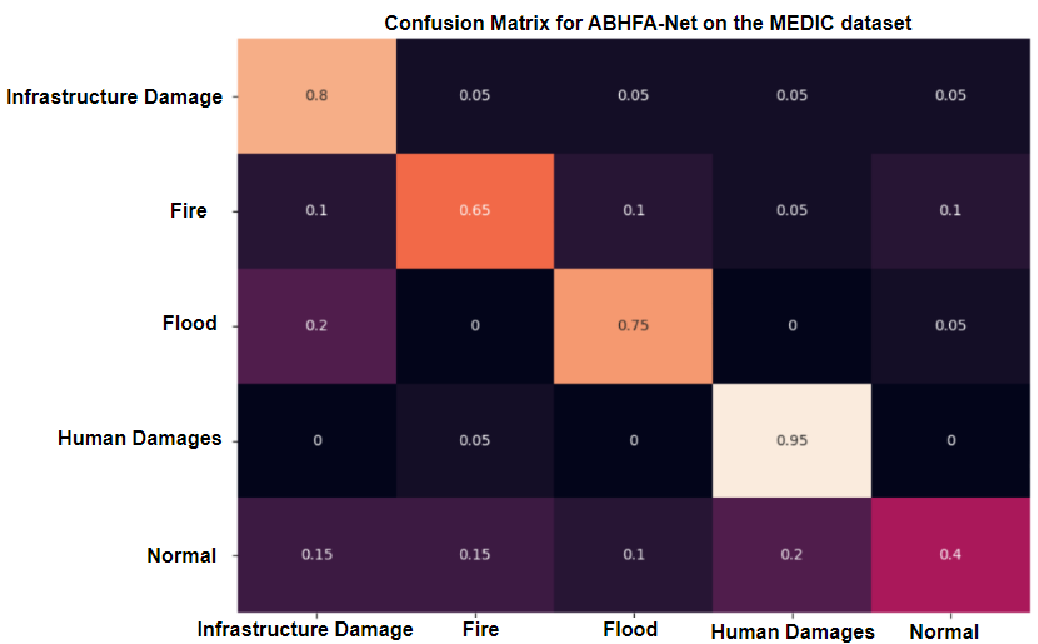}
    \caption{The confusion matrix generated for the 5-way-5-shot approach for our ABHFA-Net on the MEDIC datasets.}
\label{fig:confused_matrix_MEDIC}
\end{figure}

\begin{figure}[hbt!]
    \centering
    \includegraphics[scale=0.6]{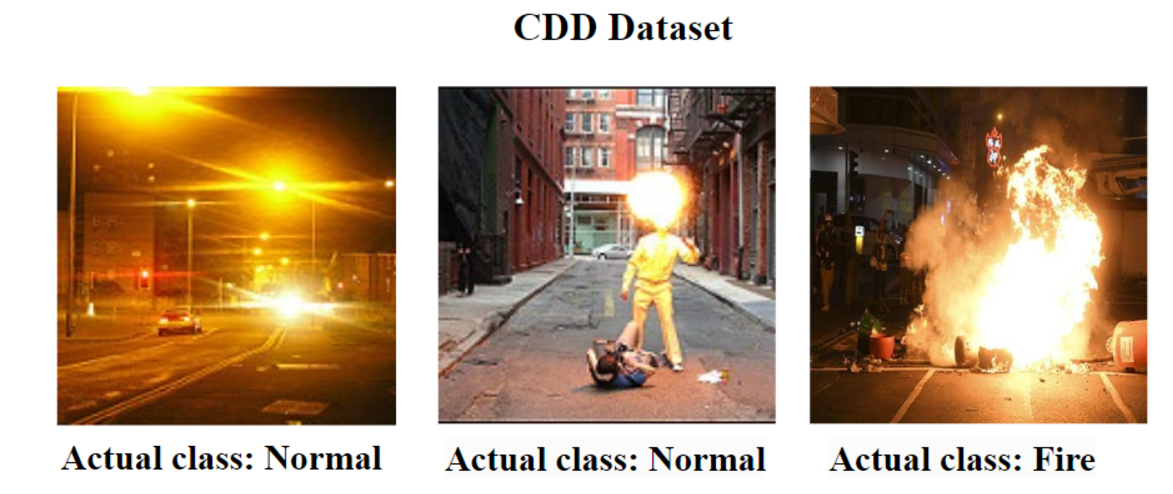}
    \caption{Example of a possible class confusion that might arose when utilizing the CDD dataset due to the wider context available. The image on the left and middle is easily confused by our algorithm for fire disaster due to its prominent orange streak and light. The image on the far right depicts an actual fire class for comparison.}
    \label{fig:confused_class_CDD}
\end{figure}

\begin{figure}[hbt!]
    \centering
    \includegraphics[scale=0.6]{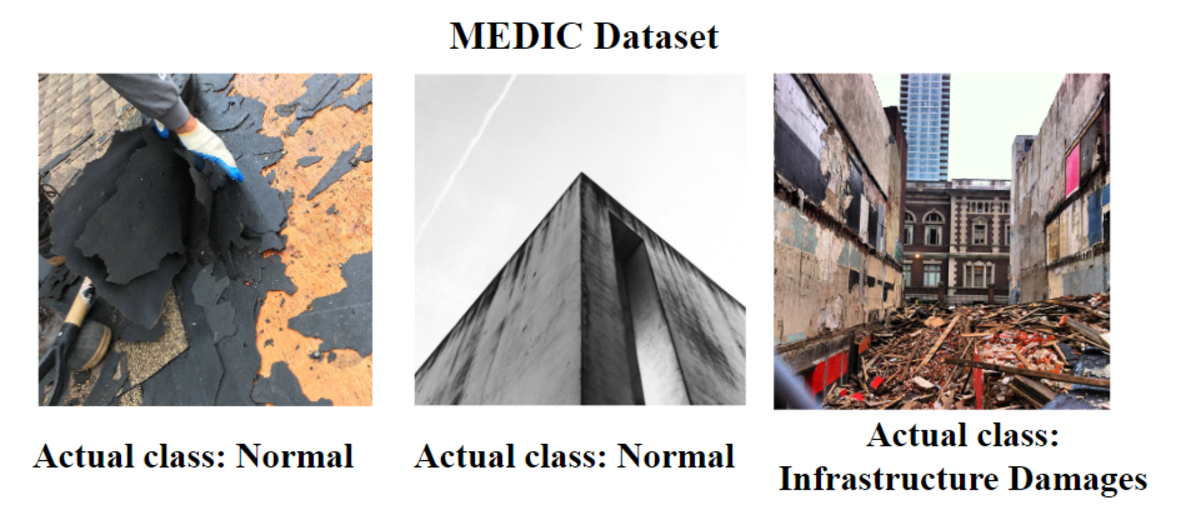}
    \caption{Example of a possible class confusion that might arose when utilizing the MEDIC dataset due to the wider context available. The image on the left and middle is easily confused by our algorithm for the damaged infrastructure class due to the relevant visual signature present (e.g., peeled and worn-off wall). The image on the far right depicts an actual damaged infrastructure class for comparison.}
    \label{fig:confused_class_MEDIC}
\end{figure}

The generally lower AA and F1-scores obtained from the methods evaluated on the CDD and MEDIC datasets, relative to that of the AIDER, can be explained, again, on the basis of the wider variety of contexts displayed for a certain disaster class in CDD and MEDIC. As mentioned earlier, we assigned certain imagery set of certain sub-classes into one class due to visual similarities present in some of the images. However, such assumptions do not lie for some CDD and MEDIC image contexts. To elaborate on this, we analyze and generate confusion matrices for the aforementioned two datasets, with Fig.\ref{fig:confused_matrix_CDD2} depicting that for the 5-way-5-shot evaluation of the CDD, and Fig.\ref{fig:confused_matrix_MEDIC} depicting the corresponding for the MEDIC. We can see that in Fig.\ref{fig:confused_matrix_CDD2}, the classes which have major misclassification is of the human damages and normal classes, as they yielded lower prediction probability of 85.0$\%$ and 65.0$\%$ respectively. The normal classes hence has the lowest prediction probability and consequently the highest misclassification rate. This means that the images presented in the normal class contained certain visual features easily mistaken for the other disaster classes, for example the normal class could be easily mistaken for the fire class due to the orange light and ambient which might be indicative of heat source, as illustrated in Fig.\ref{fig:confused_class_CDD}. The classes that has the highest correct prediction probability are the infrastructure damages, fire and water disaster class, each having prediction probability of 90.0$\%$.



In a similar manner for Fig.\ref{fig:confused_matrix_MEDIC}, it has been that the class which has the most number of misclassification is again of the normal (non-disaster) type, with only correct prediction probability of 40.0$\%$. The misclassification now ranges not only to the fire classes but also to the infrastructure damage class, which is justified by looking at Fig.\ref{fig:confused_class_MEDIC} that depicts normal class images that are easily mistaken for damaged infrastructure. The image on the left shows a peel-off surface, which can be mistaken as a possible visual signature in damaged infrastructure, and thus the misclassification, while the image on the middle shows a building edge which has some elements of wearing out, which is another possible signature of a damaged infrastructure. The image on the far right depicts an image of an actual damaged infrastructure class for reference. For the rest of the class predicted by our ABHFA-Net in MEDIC, the human damage class has the highest correct prediction probability of 95.0$\%$, followed by the damaged infrastructure, which has a correct prediction probability of 80.0$\%$. The flood class has a correct prediction probability of 75.0$\%$, while the fire class has a correct prediction probability of 65.0$\%$.

The comparatively lower class-wise accuracy observed for the normal and fire categories in MEDIC suggests that these classes exhibit substantial visual overlap with the other disaster categories. This limitation is inherent to few-shot prototype estimation under high intra-class variance and motivates future investigation into adaptive prototype calibration and class-balanced episodic learning. MEDIC is intentionally more challenging than AIDER and CDD because it contains greater semantic diversity and stronger inter-class overlap. Therefore, class-wise degradation is expected for all few-shot methods, and ABHFA-Net should be evaluated primarily through its relative improvement over competing baselines.


Lastly, it is also worthwhile to mention that some work such as \cite{he2022generating} and \cite{mirik2012utility} has emphasized the importance of utilizing a finer spatial resolution during the classification stages and assessing its impact on the classification performance. The significance of this issue is put forth since in disaster scenarios, the signature features of such disasters typically only occurred within a small region in a bigger captured field of view in the imagery. Although this aspect was not thoroughly explored in this paper as our emphasis was on assessing the algorithmic effectiveness of our few-shot approaches, it is nevertheless an interesting and essential perspective that we will investigate in our future works.

\subsubsection{Error Analysis on CDD and MEDIC Datasets}

To better understand the observed performance degradation on the CDD and MEDIC datasets relative to the AIDER dataset, we conducted a detailed error analysis based on the confusion matrices. We identified three key issues: (i) The more significant class imbalance effect, (ii) Higher intra-class variability, (iii) More frequent semantic overlaps between inter-classes.

Firstly, both CDD and MEDIC datasets exhibit significant class imbalance, in which the non-disaster (normal) class dominates the dataset more than that of AIDER (e.g., CDD has over 9000 normal images vs $<$ 1500 for other classes). This is also due to the manner in which we categorized all supposedly normal class images and labels in one class, even though in the original datasets such class are further categorized into sub-classes (e.g, CDD has streets, forest, human and sea, all with non-disaster origin). This imbalance biases the model toward predicting the majority class, leading to increased false positives and reduced sensitivity to minority disaster classes.

Secondly, as mentioned and illustrated (see Fig.\ref{fig:confused_matrix_CDD2}, \ref{fig:confused_matrix_MEDIC}, \ref{fig:confused_class_CDD} and \ref{fig:confused_class_MEDIC}) in the previous section, the CDD and MEDIC datasets contain more highly diverse visual representations within the same class (i.e., higher intra-class variability) than that of AIDER. This high variance further reduces the discriminative power of prototype-based methods even under distribution modeling.

Thirdly, the more frequent semantic overlaps encountered between normal and disaster-related images in CDD and MEDIC due to the larger and wider variety of non-disaster images (as compared to AIDER) leads to further confusion between the normal and  disaster classes. Under the 5-way-1-shot setting, limited support examples further exacerbate these issues, resulting in unstable prototype estimation and increased misclassification rates.

The first challenge involving the severe class-imbalance can be mitigated via incorporating class-balanced loss functions or utilizing re-weighting strategies, such as focal loss or inverse-frequency weighting, to penalize misclassification of minority disaster classes more strongly. The second challenge involving the higher intra-class variability could be mitigated via incorporating hard negative mining, which helps improve intra-class feature discrimination. By explicitly sampling visually similar normal images during training, the model can learn more robust decision boundaries. Lastly, as ABHFA-Net relies on distribution comparison, an additional regularization term that enforces inter-class distribution separation (e.g., margin based Bhattacharyya loss) could help reduce the overlaps between normal and disaster classes. \\

\subsection{GRAD-CAM} 

\begin{figure}[hbt!]
    \centering
    \includegraphics[scale=0.76]{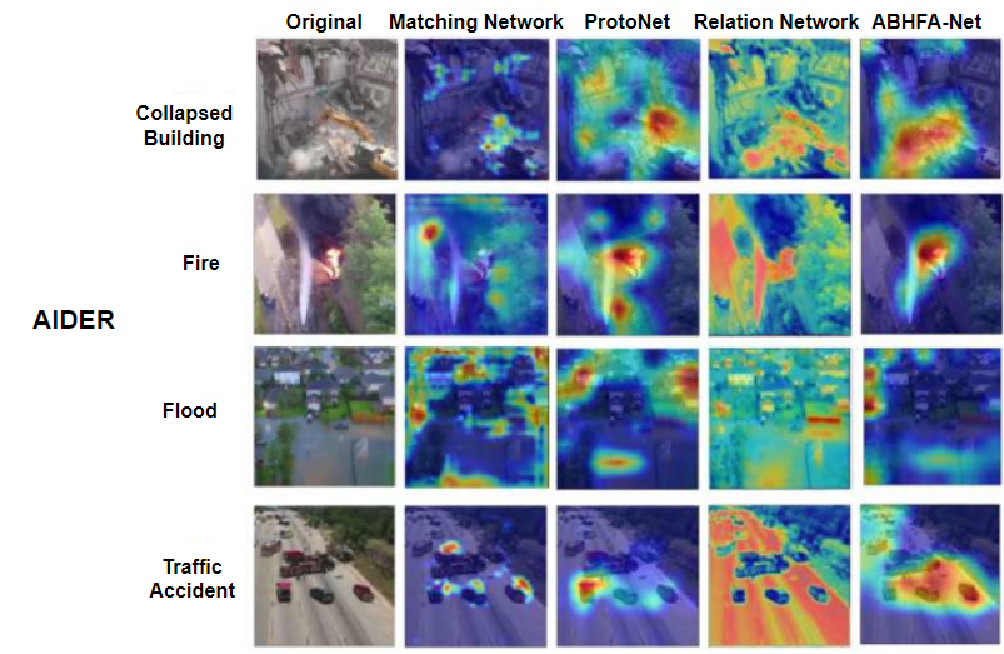}
    \caption{The grad-cam attention maps for selected FSL approaches on the AIDER images for each of the disaster classes. We excluded the class of non-disaster (normal) images in the analysis.}
    \label{fig:AIDER_gradcam_ALL}
\end{figure}

\begin{figure}[hbt!]
    \centering
    \includegraphics[scale=0.63]{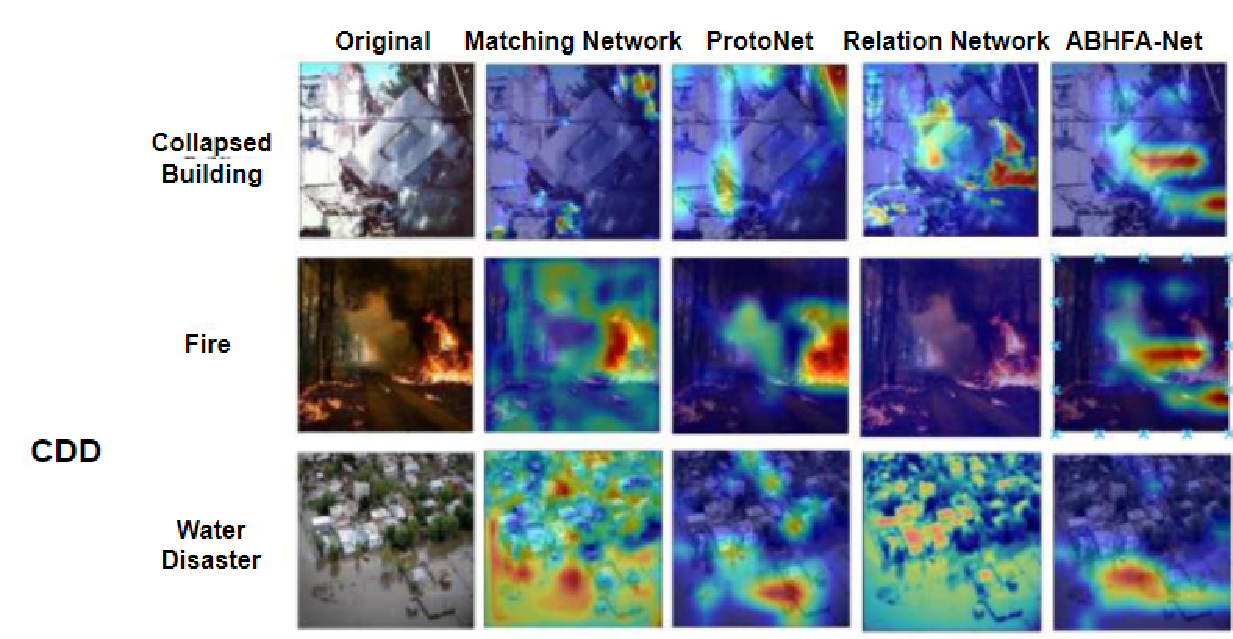}
    \caption{The grad-cam attention maps for selected FSL approaches on the CDD images for each of the disaster classes. Like AIDER, we excluded the non-disaster (normal) image class in the analysis.}
    \label{fig:CDD_gradcam_ALL}
\end{figure}

\begin{figure}[hbt!]
    \centering
    \includegraphics[scale=0.76]{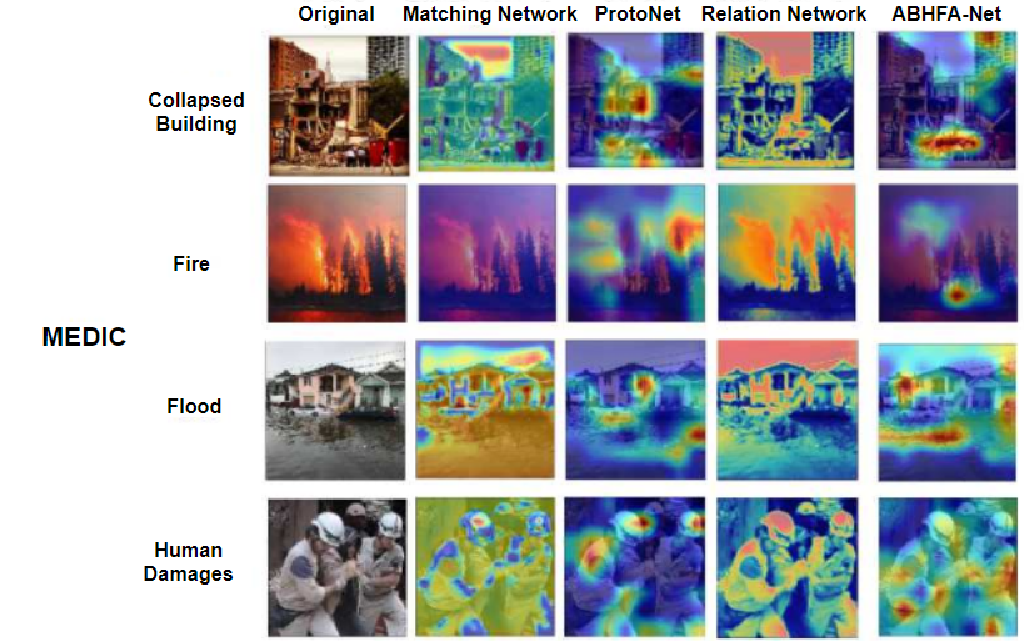}
    \caption{The grad-cam attention maps for selected FSL approaches on the MEDIC images for each of the disaster classes. Like AIDER and CDD, We excluded the non-disaster (normal) images class in the analysis.}
    \label{fig:MEDIC_gradcam_ALL}
\end{figure}

We also highlighted some of the Gradient-weighted Class Activation Mapping (GRAD-CAM) activation map on selected AIDER disaster images Fig. \ref{fig:AIDER_gradcam_ALL}, CDD in Fig.\ref{fig:CDD_gradcam_ALL}, and MEDIC in Fig.\ref{fig:MEDIC_gradcam_ALL}. Using such maps, we can dive into how our ABHFA-Net interprets the respective disaster classes in the respective images in the datasets compared to the selected SOTAs (MatchingNet, RelationNet, and ProtoNet). 

For the AIDER collapsed building image, the key feature that represents the class is the rubble near the orange bulldozer. As we can see in Fig. \ref{fig:AIDER_gradcam_ALL}, our ABHFA-Net successfully picks up the cue at that region on a broader scale, while the other SOTAs either pick it up on a smaller region (i.e., MatchingNet, RelationNet) or pay attention to the wrong part of the region (i.e., ProtoNet). Likewise, for the fire setting, our ABHFA-Net is able to pinpoint the fire hotspot located at the center accurately, whereas for the MatchingNet, the heat map pinpoints the wrong part of the fire image, while the ProtoNet and the RelationNet also pinpoint elevant areas apart from the hotspot. A more challenging scenario emerged from the flood image. As we can observe, the flooded areas are more conspicuous at the bottom of the image. For all of the selected methods, the heatmap is only able to focus on the region partially. The reason may be attributed to the grayish contrast of the flooded features, which can be easily confused for the background houses with gray rooftops.

For the CDD GRAD-CAM analysis, we once again found that most of the selected algorithm pinpoints the fire hotspot (to the right of the fire image as in Fig.\ref{fig:CDD_gradcam_ALL}) accurately, with MatchingNet and ProtoNet being able to highlight the spot entirely. For ABHFA-Net, it seems that the attention maps is more spread out than the two aforementioned methods, while the RelationNet fails to spot the heat region. Our ABHFA-Net, instead, in this context is able to pinpoint the key flood regions and building rubble regions more conspicuously than the other methods. Lastly, for the GRAD-CAM analysis of the MEDIC images, our ABHFA-Net once again successfully highlights the key flood regions for the flood image, along with the key rubble regions in the collapsed building image, as in Fig \ref{fig:MEDIC_gradcam_ALL}. However, we note that for the latter, the ProtoNet is also able to point out the exposed component of the remaining, standing part of the building, which is also a hallmark feature or signs of a building destruction. For the forest fire image, ProtoNet and RelationNet are able to successfully highlight a larger region of the smokes and fire hotspot on top of the tall trees, while our ABHFA-Net pinpoints only the bottom part of the hotspot, which is not as conspicuous as the top part, and MatchingNet fails to detect any key region at all (like that of the RelationNet in the fire image in the CDD). 


An interesting point of discussion lies on the image classified under human damages, which depicts a lady being rescued and led out by two rescue personnel. Such image actually highlighted visually the limitations of the current methods as well as our approach. This is because to interpret the image to be of some kind of human distress, the overall context has to be taken into account, as the definition of human distress is more complicated and takes on a wide variety of forms and appearances. According to \cite{singh2022transductive}, such context attributes would constitute the \emph{contextual} information, which separates it from the \emph{label} information that contains more class-characterizing attributes. Although the author mentioned that the contextual information may not be relevant characteristics that distinguishes images of different classes, instead relying on the label information, this may only be true for the other disaster classes (food, fire, collapsed building, traffic accident) in our setting which always contains outstanding attributes that warrant immediate distinction regardless of its forms and appearances, as we have discussed in previous and current GRAD-CAM analysis. Therefore, as we can see in the last row of Fig.\ref{fig:MEDIC_gradcam_ALL}, merely highlighting the helmet of the rescuers and the key body pose, as is done for our ABHFA-Net and RelationNet, may not be sufficient for accurately depicting the correct class, which means that the heat map generated needs to display higher generalization and robustness in the form of a wider-scale map distribution. This issue can primarily contribute and explain the reason behind the lower AA values obtained for the MEDIC dataset with respect to CDD and AIDER, regardless of the approaches utilized. 

In summary, the above GRAD-CAM analysis indicated that our ABHFA-Net has correctly highlighted the key visual signature for each disaster types in AIDER to a high level of degree, while yielding mixed results for the other two datasets. Although the analysis was done for one image per disaster class for all the datasets, the observations and trends can be extrapolated to the rest of the images, since we have mentioned beforehand that the CDD and MEDIC have more disaster feature variations and contexts than AIDER, which would lead to higher probability of wrong GRAD-CAM highlights regardless of the approaches. This may help to partially explain why ABHFA-Net has the highest AA for both the 1-shot and 5-shot approaches in AIDER, and why the overall AA values are lower for the other SOTAs for CDD and MEDIC relative to AIDER.

\subsection{Ablation Study} 

We conducted an ablation study investigating the role of each of our ABHFA-Net's novel component on 1-shot and 5-shot performance using the three dataset. The tabulated results are displayed in Table \ref{tab:Ablation_AIDER}-\ref{tab:Ablation_MEDIC}. For a more holistic evaluation, we also compared our model's performances when we replaced the $\ell_{BHAS}$ with the KL divergence loss $\ell_{KL}$, the Wasserstein Loss $\ell_{Wasserstein}$, as well as the Hellinger Similarity Loss $\ell_{Hess}$ proposed in HELA-VFA. Take note that throughout the tables, the configuration labeled ``W/o $\ell_{BHAS}$, $\ell_{Hess}$, $\ell_{Wasserstein}$ or $\ell_{KL}$'' and ``W/o att, $\ell_{BHAS}$, $\ell_{KL}$, $\ell_{Wasserstein}$ or $\ell_{Hess}$'' denote ABHFA-Net training using only the cross-entropy loss. It is also of interest to note that the configuration labeled ``Attention + $\ell_{Hess}$'' actually represents the HELA-VFA configuration.

We can observe from the tables that relatively lower values are obtained for all shot setting in all selected datasets when attention was not included, regardless of whether $\ell_{KL}$, $\ell_{Wasserstein}$, $\ell_{Hess}$, or $\ell_{BHAS}$ is used (or when all of them are not used). We also noted that, regardless or not attention is included, the classification values that are obtained using $\ell_{KL}$ and $\ell_{wasserstein}$ are in general lesser than those of $\ell_{Hess}$ or $\ell_{BHAS}$. This can be attributed to the Bhattacharyya distance being a more generalized distance metric than that of the KL divergence and Hellinger distance due to its additional capability to model the covariance structure for reasons already discussed in the previous sections. Furthermore, due to the wider variety of classes presented in the CDD and MEDIC dataset, as compared to AIDER, the drop in the classification accuracy values becomes more apparent when essential components of the network are removed, regardless of whether $\ell_{KL}$, $\ell_{Wasserstein}$, $\ell_{Hess}$ or $\ell_{BHAS}$ is utilized. This is more pronounced for the MEDIC dataset, whereby the drop in performances for both shot settings between the configuration ``W/o $\ell_{BHAS}$, $\ell_{Hess}$, $\ell_{Wasserstein}$ or $\ell_{KL}$" and ``W/o att, $\ell_{BHAS}$, $\ell_{Hess}$, $\ell_{Wasserstein}$ or $\ell_{KL}$'' is more than 5$\%$. This also quantitatively illustrate the role of attention in focusing on higher-level contextual features to generalize performances to unseen data, and once again justified the inclusion of attention into our network, irrespective of whether $\ell_{BHAS}$, $\ell_{Hess}$, $\ell_{Wasserstein}$, or $\ell_{KL}$ is used.

\begin{table}
\caption{Ablation study results on the ABHFA-Net for the various configurations tabulated below on the AIDER dataset. Bold values denote the optimal values obtained out of all the configurations attempted.}
\begin{tabular}{p{6cm}p{3cm}p{3cm}}
\hline
\textbf{Configuration} & \textbf{5-way-1-shot} & \textbf{5-way-5-shot} \\
\hline
\textbf{Attention + $\ell_{BHAS}$} & \textbf{68.2}$\pm$\textbf{1.19} & \textbf{78.3}$\pm$\textbf{0.75} \\
No Attention + $\ell_{BHAS}$ & 65.5$\pm$0.37 & 76.2$\pm$0.54\\
Attention + $\ell_{Hess}$ & 67.7$\pm$0.78 & 75.7$\pm$0.98 \\
No Attention + $\ell_{Hess}$ & 67.0$\pm$0.61 & 75.4$\pm$0.25\\
No Attention + $\ell_{Wasserstein}$ & 62.7$\pm$0.99 & 66.1$\pm$1.16\\
Attention + $\ell_{KL}$ & 64.5$\pm$0.55 & 75.9$\pm$0.23 \\
No Attention + $\ell_{KL}$ & 62.7$\pm$0.36 & 75.4$\pm$0.96 \\
W/o $\ell_{BHAS}$, $\ell_{Hess}$, $\ell_{Wasserstein}$ or $\ell_{KL}$ & 62.0$\pm$0.67 & 73.2$\pm$0.85\\
W/o att, $\ell_{BHAS}$, $\ell_{Hess}$, $\ell_{Wasserstein}$ or $\ell_{KL}$ & 60.4$\pm$0.29 & 71.4$\pm$0.21\\
\hline
\end{tabular}
\label{tab:Ablation_AIDER}
\end{table}

\begin{table}
\caption{Ablation study results on the ABHFA-Net for the various configurations tabulated below on the CDD dataset. Bold values denote the optimal values obtained out of all the configurations attempted.}
\begin{tabular}{p{6cm}p{3cm}p{3cm}}
\hline
\textbf{Configuration} & \textbf{5-way-1-shot} & \textbf{5-way-5-shot} \\
\hline
\textbf{Attention + $\ell_{BHAS}$} & \textbf{65.0}$\pm$\textbf{0.95} & \textbf{74.2}$\pm$\textbf{0.65} \\
No Attention + $\ell_{BHAS}$ & 61.2$\pm$1.77 & 73.6$\pm$1.25\\
Attention + $\ell_{Hess}$ & 64.3$\pm$0.73 & 73.0$\pm$0.51 \\
No Attention + $\ell_{Hess}$ & 59.8$\pm$0.47 & 72.3$\pm$0.53 \\
No Attention + $\ell_{Wasserstein}$ & 61.2$\pm$1.18 & 65.8$\pm$1.08\\
Attention + $\ell_{KL}$ & 60.1$\pm$0.44 & 71.5$\pm$0.95  \\
No Attention + $\ell_{KL}$ & 58.6$\pm$0.79 & 70.1$\pm$0.47 \\
W/o $\ell_{BHAS}$, $\ell_{Hess}$, $\ell_{Wasserstein}$ or $\ell_{KL}$ & 56.0$\pm$1.27 & 68.1$\pm$0.53\\
W/o att, $\ell_{BHAS}$, $\ell_{Hess}$, $\ell_{Wasserstein}$ or $\ell_{KL}$ & 53.3$\pm$0.78& 66.7$\pm$1.66\\
\hline
\end{tabular}
\label{tab:Ablation_CDD}
\end{table}

\begin{table}
\caption{Ablation study results on the ABHFA-Net for the various configurations tabulated below on the MEDIC dataset. Bold values denote the optimal values obtained out of all the configurations attempted.}
\begin{tabular}{p{6cm}p{3cm}p{3cm}}
\hline
\textbf{Configuration} & \textbf{5-way-1-shot} & \textbf{5-way-5-shot} \\
\hline
\textbf{Attention + $\ell_{BHAS}$} & \textbf{60.2}$\pm$\textbf{1.22} & \textbf{66.5}$\pm$\textbf{0.95} \\
No Attention + $\ell_{BHAS}$ & 58.5$\pm$0.56 & 64.9$\pm$1.43\\
Attention + $\ell_{Hess}$ & 58.4$\pm$0.96 & 66.1$\pm$1.56 \\
No Attention + $\ell_{Hess}$ & 55.1$\pm$1.41 & 65.2$\pm$1.34 \\
No Attention + $\ell_{Wasserstein}$ & 55.9$\pm$1.09 & 59.8$\pm$1.68\\
Attention + $\ell_{KL}$ & 57.7$\pm$1.77 & 65.2$\pm$0.71  \\
No Attention + $\ell_{KL}$ & 52.3$\pm$1.34 & 59.8$\pm$1.70 \\
W/o $\ell_{BHAS}$, $\ell_{Hess}$, $\ell_{Wasserstein}$ or $\ell_{KL}$ & 48.7$\pm$0.83 & 54.1$\pm$1.56\\
W/o att, $\ell_{BHAS}$, $\ell_{Hess}$, $\ell_{Wasserstein}$ or $\ell_{KL}$ & 42.2$\pm$1.02 & 48.2$\pm$0.75 \\
\hline
\end{tabular}
\label{tab:Ablation_MEDIC}
\end{table}

To investigate the effect of attention design in ABHFA-Net, we conducted additional ablation study in which we replaced the proposed channel-spatial attention module with alternative attention mechanisms commonly used in FSL, including transformer self-attention (Vaswani et al. \cite{vaswani2017attention}), cross-attention (Xiao et al. \cite{xiao2022semantic}), attention with weight fusion (Meng et al. \cite{meng2023few}), as well as the attention utilized in Squeeze and Excite (SE) blocks \cite{hu2018squeeze} and ECA-Net (Wang et al. \cite{wang2020eca}), and assessed their effect on the classification AA and efficiency in terms of FLoating Point Operations per second (FLOPs in $\times 10^{6}$). The results are illustrated in Table \ref{tab:Ablation_Attention_AIDER} for AIDER. For a fair comparison, all variants use the same ResNet12 backbone, episodic sampling protocol, Bhattacharyya-distance-based classifier, and loss function. Only the attention module inside the encoder is changed. Incorporating the channel-spatial attention mechanism has been shown to greatly reduced the FLOPs value relative to the others (but still required higher FLOPs than without any attention by 2$\%$), although cross-attention yielded slightly higher accuracy scores than the channel-spatial attention, at the expense of significantly higher FLOPs requirement. This suggests that jointly emphasizing informative feature channels and spatial disaster regions is particularly effective for distribution-based prototype learning, unlike self-attention or cross-attention which introduce stronger pairwise interactions and higher computational cost. Channel-spatial attention also provides a lightweight refinement of the encoder features before the latent Gaussian parameters estimation procedure, leading to more stable prototype distributions and improved variational classification. Lastly, the aforementioned attention also surpassed that of the ECA-Net and SE blocks, as the latter two only model channel importance, whereas in few-shot distribution estimation both channel and spatial aspects of the visual features are crucial.

\begin{table}
\caption{Ablation study results on the ABHFA-Net for the various attention mechanism tabulated below on the AIDER dataset. In addition to the AA values, efficiency metrics such as FLOPs (in $\times 10^{6}$) was also included. The ``SE attention, WF attention, and CS attention" represents the Squeeze and Excite, Weight-Fusion and Channel-Spatial attention respectively. Bold values denote the optimal values obtained out of all the configurations attempted.}
\begin{tabular}{p{4cm}p{2cm}p{2cm}p{2cm}}
\hline
\textbf{Configuration} & \textbf{5-way-1-shot} & \textbf{5-way-5-shot} & \textbf{FLOPs ($\times 10^{6}$)} \\
\hline
No Attention & 65.5$\pm$0.37 & 75.2$\pm$0.54 & \textbf{65.0} \\
Self-Attention \cite{vaswani2017attention} & 67.3$\pm$0.91 & 78.0$\pm$0.88 & 81.0  \\
\textbf{Cross-Attention \cite{xiao2022semantic}} & \textbf{68.9$\pm$1.05} & \textbf{79.5$\pm$0.92} & 91.0 \\
WF Attention \cite{meng2023few} & 67.0$\pm$0.80 & 77.5$\pm$0.73 & 69.5 \\
SE Attention \cite{hu2018squeeze} & 66.4$\pm$0.65 & 76.7$\pm$0.58 & 66.0 \\
ECA-Net \cite{wang2020eca} & 66.7$\pm$0.52 & 76.9$\pm$0.57 & 65.2 \\
CS Attention & 68.2$\pm$1.19 & 78.3$\pm$0.75 & 67.0 \\
\hline
\end{tabular}
\label{tab:Ablation_Attention_AIDER}
\end{table}

\subsection{t-SNE}

\begin{figure}[hbt!]
    \centering
    \includegraphics[scale=0.60]{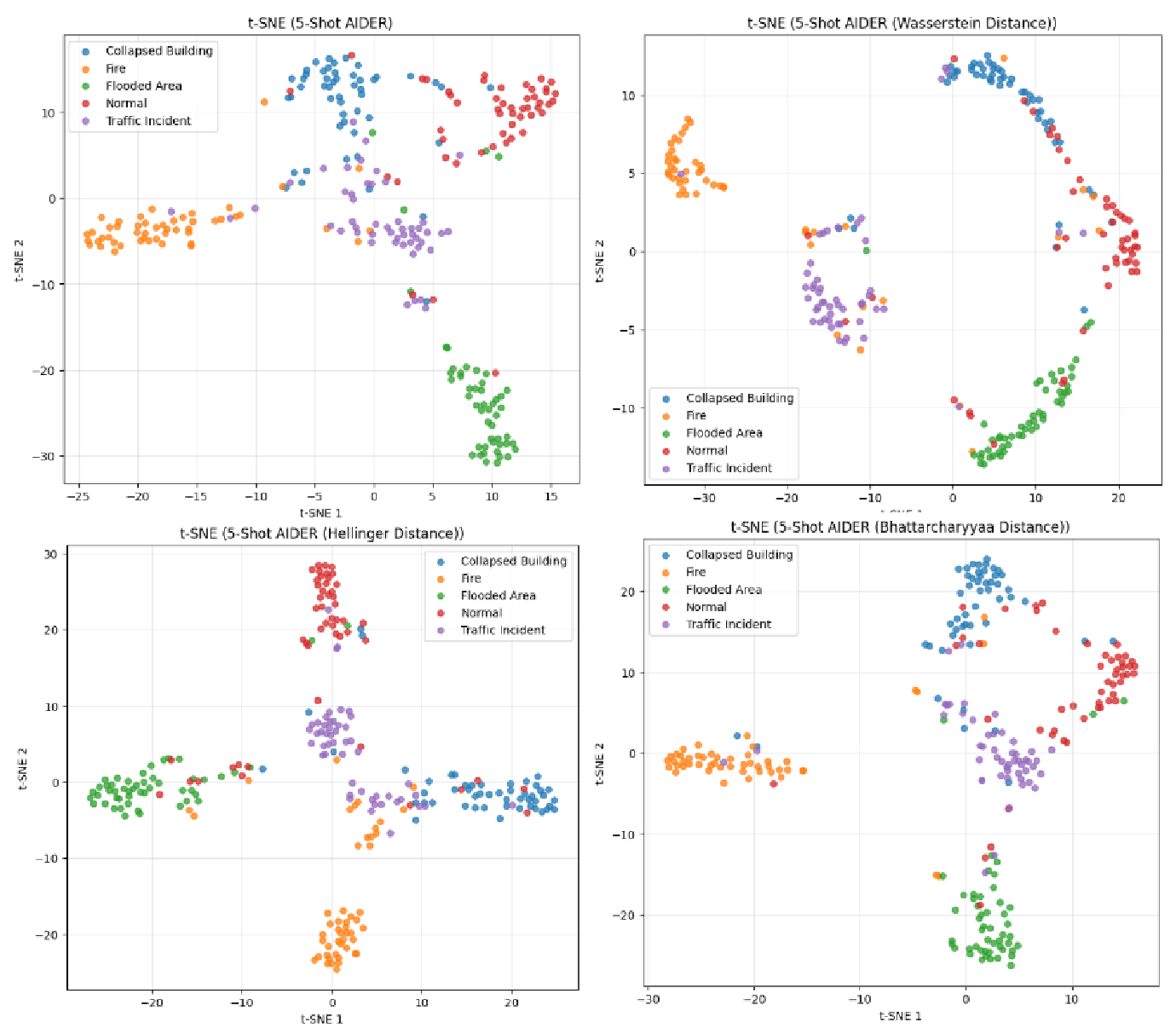}
    \caption{t-SNE plots for our variational model but utilizing different distances. \textbf{(Top Left):} Using the KL divergence. \textbf{(Top Right):} Using the Wasserstein distance. \textbf{(Bottom Left):} Using the Hellinger distance. \textbf{(Bottom Right):} Using the Bhattarcharyya distance.}
    \label{fig:tSNE_Compare}
\end{figure}

We have also provided a qualitative visualization of the feature embedding capabilities of our ABHFA-Net via t-SNE plots and illustrate how our ABHFA-Net distribution modeling can enhance class separation relative to other distance metrics. These are illustrated in Fig.\ref{fig:tSNE_Compare} for the case of the KL divergence, Wasserstein distance, Hellinger distance and the Bhattarcharyya distance, as applied to 5 class clusters of AIDER.

The t-SNE visualizations reveal distinct structural differences in the latent representations induced by different distance metrics. The KL divergence results in scattered and partially overlapping clusters, reflecting its asymmetric nature and lack of explicit separation constraints. The Wasserstein distance produces well-separated but elongated clusters, preserving global geometry but lacking compactness, as it emphasizes more on geometric alignment between distributions. The Hellinger distance yields more balanced representations with moderate compactness and separation. In contrast, Bhattacharyya distance achieves both strong intra-class compactness and clear inter-class separation. This is due to its explicit consideration of  both mean differences and covariance overlap, leading to more discriminative  and stable class distributions. These properties are particularly beneficial in few-shot settings, where robust prototype estimation is critical.




\subsection{Further Limitations}

A few other limitations still exist apart from those discussed in the deep error analysis subsection.

Firstly, our variational formulation assumes Gaussian distributions for both the latent prior and the approximate posterior. Such assumption is standard in variational latent variable models and serves as a tractable and reasonable approximation under limited data. However, it may not fully capture the complex multi-modal or heavily tailed latent distribution structures arising in highly heterogeneous disaster scenery.

Secondly, our ABHFA-Net incorporates spatial–channel attention modules to enhance the discriminative feature aggregation. Although these proposed modules introduced additional computational overhead relative to the vanilla convolutional encoders, the overhead remains moderate relative to the overall network cost and is significantly lower than that of deeper backbone architectures. Nevertheless, in resource-constrained environment, for instance the onboard UAV operations and inference, this overhead may become non-negligible.

Lastly, like most metric-and prototypical-based few-shot learning approaches, ABHFA-Net may remain sensitive to outliers or mislabeled samples in the support and query sets, particularly under extremely low-shot conditions. Although modeling class prototypes as distributions rather than point estimates could improves robustness to intra-class variance, the extreme outliers can still bias the estimated latent statistics.

For the first limitation, while mixtures of Gaussians can partially alleviate the challenge, future work will explore more expressive posterior families, such as normalizing flows or implicit variational distributions, to relax the Gaussian assumption while retaining tractable inference. As for the second limitation, future work would involve exploring more lightweight attention variants and dynamic attention activation strategies to further reduce computational cost while not compromising the classification performance. Lastly, for the third limitation, works that could be explored include incorporating robust estimation techniques, uncertainty-aware sample weighting, or outlier detection mechanisms within the episodic training framework. 

\section{Additional UAV Disaster Robustness Evaluation}

\begin{figure}[hbt!]
    \centering
    \includegraphics[scale=0.38]{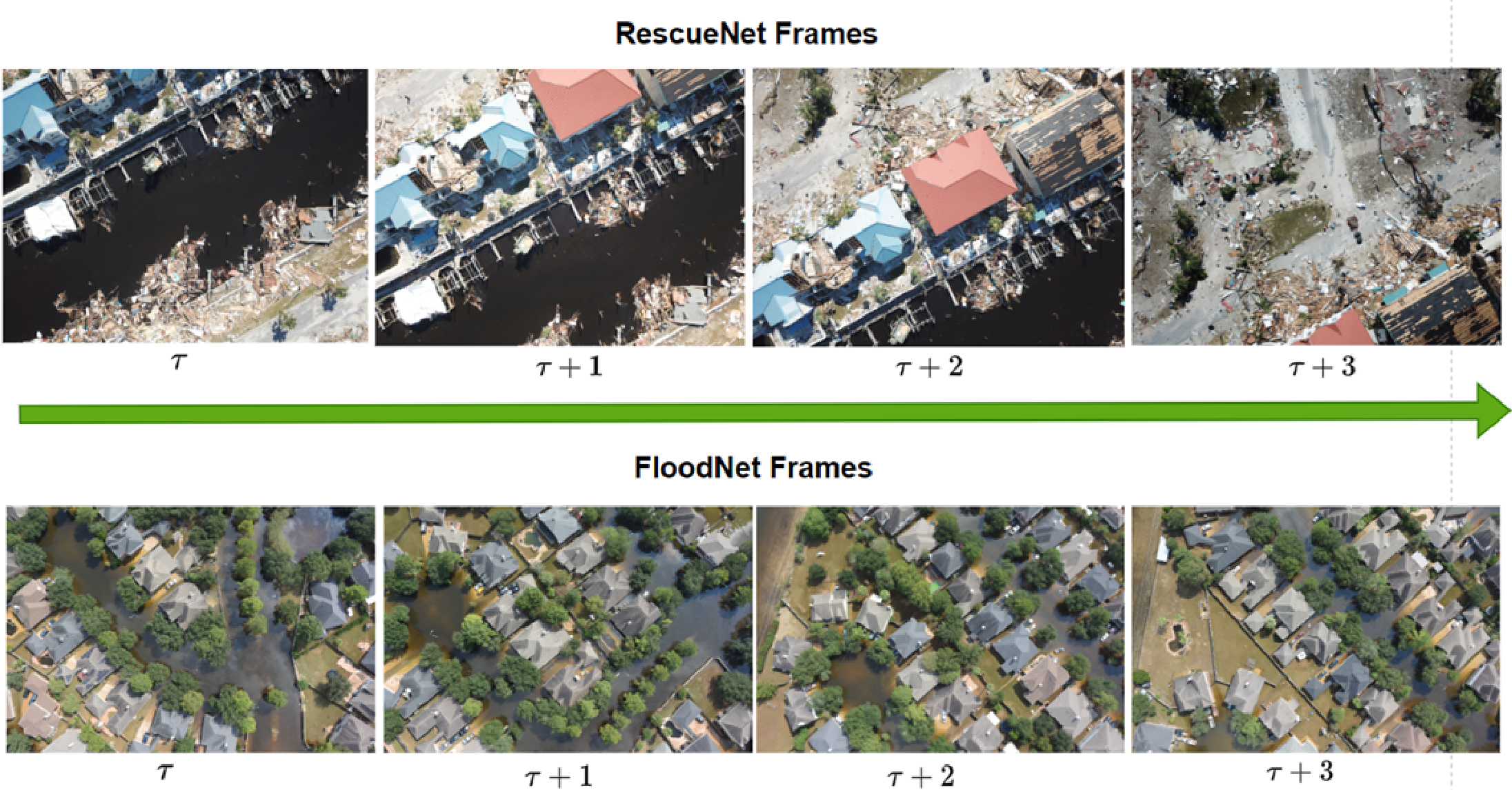}
    \caption{Sequential UAV frames illustration for RescueNet (\textbf{Top}) and FloodNet (\textbf{Bottom}) for frame $\tau$, $\tau+1$, $\tau+2$ and $\tau+4$.}
    \label{fig:FloodNet_RescueNet}
\end{figure}

Although a full onboard UAV deployment was beyond the scope of the present study, we conducted a deployment-oriented simulation using RescueNet \cite{rahnemoonfar2023rescuenet} and FloodNet \cite{rahnemoonfar2021floodnet} frames, which also serves as a cross-domain evaluation to analyze the generalization capability of our ABHFA-Net. These two datasets are both UAV-based post-disaster datasets: RescueNet contains high-resolution UAV imagery after Hurricane Michael with detailed damage annotations, while FloodNet contains UAV flood-scene imagery after Hurricane Harvey and supports image classification, segmentation, and Visual-Question-Answer (VQA) tasks. The images involved in FloodNet and RescueNet were both collected and generated from UAV surveys conducted in real disaster environments as high-resolution still photographs, unlike the AIDER/CDD/MEDIC which involved the myriad of disaster and non-disaster scenery collected either from UAV or from google images. Therefore the FloodNet and RescueNet evaluation is performed on imagery acquired under more realistic aerial observation conditions, allowing noisy image/low-resolution robustness, and inference latency/Frame-Per-Second (FPS) to be tested. Consequently, they offer unseen disaster perspectives and contexts rather than merely serving as additional test images, allowing assessment of the robustness of the learned few-shot representations under realistic deployment scenarios. Some examples of the sequential UAV frames in the RescueNet and FloodNet are illustrated in Fig.\ref{fig:FloodNet_RescueNet}. For both datasets, the images were collected using a DJI Mavic Pro quadcopter that performs over several individual flights to individual affected and unaffected areas.

Unlike conventional few-shot image classification benchmarks (as in previous sections) that provide a single categorical label for each image, both RescueNet and FloodNet provide pixel-level semantic ground-truth maps, where each pixel is assigned a semantic class (e.g., flooded road and building, water, vegetation, background). These dense annotations provide substantially richer supervisory information by explicitly describing the spatial distribution and extent of flooding within each UAV image rather than indicating only image-level category. To enable evaluation via ABHFA-Net, which is initially formulated as a few-shot image classification framework, the pixel-level ground-truth maps were transformed into image-level severity labels. Specifically, the proportion of pixels belonging to the respective semantic class was computed for each annotation mask, and predefined thresholds were subsequently used to assign the image into low-, moderate-, or severe-disaster affected categories. The semantic annotations stand as the source of supervision while providing image-level labels compatible with episodic few-shot learning.

The experimental hyperparameters when evaluating our model on the two datasets are the same as that outlined in Table \ref{tab:hyperparams}, except for the training episodes (which we reduced to 10000), as well as for the image sizes (which has been scaled to 256$\times$128 from the original 3000$\times$4000 resolution per image). The reason for such scaling is similar to that of section 5.1, as the model's efficiency is hugely dependent on the image sizes. For this experiment, we do not incorporate the 95$\%$ confidence interval for a given number of trials as we intend the inference to be completed over one trial, simulating the condition of direct and immediate model inference of the disaster contexts. 

Table \ref{tab:RescueNet_FloodNet_1} depicts the AA, Macro-F1, Latency, Frame-Per-Second (FPS), peak GPU memory, and temporal consistency for the RescueNet and FloodNet evaluation using ABHFA-Net, while Table \ref{tab:RescueNet_FloodNet_2} and \ref{tab:RescueNet_FloodNet_3} depicts the AA and Macro-F1 values obtained as a function of the image resolution and Gaussian noise corruption severity per image respectively. In Table \ref{tab:RescueNet_FloodNet_1}, latency denotes the average inference time per frame, temporal consistency equates to the percentage of consecutive frames receiving identical predictions when the disaster scene is not altered significantly, and FPS is mathematically defined by $1000/Latency$, where latency is measured in millisecond (ms). In Table \ref{tab:RescueNet_FloodNet_2} we gradually scaled down the resolution of each dataset image by a factor of $\times2$ starting from 256$\times$128 to evaluate our ABHFA-Net's robustness towards resolution degradation due to increase in the altitude gain of the UAV (resulting in each object occupying lesser pixels in the field-of-view of the quadcopter). For Table \ref{tab:RescueNet_FloodNet_3}, we selected the Gaussian noise $\sigma$ values to be 0.10, 0.15, 0.20 and 0.30, in line with the work by \cite{lee2025anrot}. This simulates the imperfections in the imaging sensor and the image acquisition process during the flight (for instance, when the camera ISO is increased, the sensor induced random fluctuations (Gaussian noise) into the recorded pixel intensities).

\begin{table}
\centering
\caption{The AA, Macro-F1, Latency, Frame-Per-Second (FPS), peak GPU memory, and temporal consistency for RescueNet and FloodNet evaluation using ABHFA-Net. All images are of resolution 256$\times$128 in this study.} \label{tab:RescueNet_FloodNet_1}
\begin{tabular}{p{2cm}p{1cm}p{1cm}p{1cm}p{1cm}p{2cm}p{2cm}}
\hline
\textbf{Dataset} & \textbf{AA} & \textbf{Macro-F1} & \textbf{Latency} & \textbf{FPS} & \textbf{Peak GPU Memory} & \textbf{Temporal Consistency} \\
\hline
 RescueNet & 70.3 & 70.4 & 7.5ms/f & 140.0 & 266.4MB & 57.2$\%$ \\
 FloodNet & 96.3 & 92.4 & 11.7ms/f & 85.1 & 259.9MB & 74.7$\%$ \\
 \hline
\end{tabular}
\end{table}

From Table \ref{tab:RescueNet_FloodNet_1}, we see that the proposed ABHFA-Net achieved a mean inference latency of 7.5 ms per frame on RescueNet, corresponding to 140.0 FPS, which is 4.7$\times$ faster than a 30 FPS real-time inference requirement. Furthermore, the model required only 266.4 MB of peak GPU memory, indicating a lightweight computational footprint suitable for deployment on resource-constrained edge devices. However, the temporal consistency reached only 57.2$\%$, suggesting that predictions fluctuate considerably across consecutive frames. This lower temporal stability may be attributed to the greater scene diversity and abrupt transitions commonly observed in RescueNet, including varying viewpoints and boundaries, complex urban layouts, and occlusions, which make maintaining consistent predictions over time more challenging. In contrast, ABHFA-Net achieved substantially stronger performance on FloodNet, with an AA of 96.3$\%$ and Macro-F1 of 92.4$\%$, while maintaining a real-time inference speed of 11.7 ms per frame (85.1 FPS). Although the latency increased relative to RescueNet, the throughput remains well above the 30 FPS typically required for real-time UAV applications. Notably, the peak GPU memory is altered minimally (259.9 MB versus 266.4 MB), demonstrating consistent memory efficiency across datasets. More importantly, the temporal consistency increased from 57.2$\%$ on RescueNet to 74.7$\%$ on FloodNet, indicating that the model produced more stable predictions across consecutive frames. This improvement is likely because FloodNet contains larger and more spatially continuous flooded regions with fewer abrupt scene transitions, enabling the learned feature representations and attention mechanisms of ABHFA-Net to maintain more consistent predictions over time.

From Table \ref{tab:RescueNet_FloodNet_2}, we see an interesting trend. At moderate downsampling from 256$\times$128 to 128$\times$64, AA classification performance is improved on both RescueNet and FloodNet, from 70.3$\%$ to 76.6$\%$ and 96.3$\%$ to 98.8$\%$, respectively. This suggests that moderate resolution reduction suppresses fine-scale background textures and high-frequency noise while preserving the dominant structural characteristics of flooded regions, allowing the attention-guided prototype representations to focus more effectively on discriminative semantic features. As the resolution is further reduced to 64$\times$32 and 32$\times$16, the performance generally deteriorates in general because progressively more spatial details and object boundaries are lost, making it increasingly difficult to distinguish flood severity patterns. Nevertheless, FloodNet consistently maintains higher AA and Macro-F1 values than RescueNet across all resolutions, indicating greater robustness to spatial information loss. Again, this behaviour is likely attributable to FloodNet containing larger, more spatially coherent flooded regions, whereas RescueNet comprises more complex urban scenes with finer structural details that are more susceptible to degradation under aggressive downsampling.

\begin{table}
\centering
\caption{The RescueNet and FloodNet AAs and Macro-F1s as a function of gradual decrease in resolution by $\times2$ in the length and width of each image frames for the RescueNet and FloodNet dataset using ABHFA-Net.} \label{tab:RescueNet_FloodNet_2}
\begin{tabular}{p{2cm}p{2cm}p{2cm}p{2cm}p{2cm}}
\hline
\textbf{Resolution} & \textbf{RescueNet AA} & \textbf{FloodNet AA} & \textbf{RescueNet Macro-F1} & \textbf{FloodNet Macro-F1} \\
\hline
 256$\times$128 & 70.3 & 96.3 & 70.4 & 92.4 \\
 128$\times$64 & 76.6 & 98.8 & 76.5 & 97.3 \\
 64$\times$32 & 60.6 & 91.5 & 60.6 & 91.5 \\
 32$\times$16 & 67.4 & 83.7 & 67.4 & 83.7 \\
 \hline
\end{tabular}
\end{table}

Finally, looking at Table \ref{tab:RescueNet_FloodNet_3}, the influence of noise differs between RescueNet and FloodNet, while degradation from the previous table values are seen. On RescueNet, the AA fluctuates between 59.3$\%$ and 66.5$\%$, while the Macro-F1 varies from 39.0$\%$ to 47.5$\%$. Rather than exhibiting a monotonic decline, the performance improves at moderate corruption levels ($\sigma$ = 0.15–0.20), suggesting that mild Gaussian perturbations suppress insignificant high-frequency image variations and background clutter, thereby encouraging the attention-guided encoder to focus on more discriminative structural characteristics of flooded regions. As the corruption increases further to $\sigma$ = 0.30, both AA and Macro-F1 decrease, indicating that excessive noise eventually obscures essential spatial and texture information required for reliable prototype generation.


\begin{table}
\centering
\caption{The RescueNet and FloodNet AAs and Macro-F1s as a function of gradual Gaussian noise corruption $\sigma$ of each image frames for the RescueNet and FloodNet dataset using ABHFA-Net. All images are of resolution 256$\times$128 in this study. \label{tab:RescueNet_FloodNet_3}}
\begin{tabular}{p{2cm}p{2cm}p{2cm}p{2cm}p{2cm}}
\hline
\textbf{Corruption Severity} & \textbf{RescueNet AA} & \textbf{FloodNet AA} & \textbf{RescueNet Macro-F1} & \textbf{FloodNet Macro-F1} \\
\hline
 $\sigma = 0.10$ & 60.9 & 96.0 & 39.0 & 91.5  \\
 $\sigma = 0.15$ & 66.5 & 76.6 & 42.4 & 76.8  \\
 $\sigma = 0.20$ & 61.7 & 89.2 & 47.5 & 89.1  \\
 $\sigma = 0.30$ & 59.3 & 76.1 & 39.3 & 76.3  \\
 \hline
\end{tabular}
\end{table}


A limitation of the present study is that the pixel-level semantic annotations provided by RescueNet and FloodNet were utilized only to derive image-level flood severity labels for compatibility with our proposed classification framework. Future work will directly exploit the full semantic label maps through joint few-shot classification and semantic segmentation, enabling pixel-level flood localization while preserving the prototype-based decision mechanism. \\

\section{Conclusions}

Our approach employs UAV-based remote sensing images to tackle natural disaster scenarios. We exhibit the efficacy of ABHFA-Net in four common experimental benchmarks, three disaster classification tasks, establishing it as promising benchmarks for future investigations in this area, as well as assessing the model's practical deployment capability using the UAV-based RescueNet and FloodNet datasets. By utilizing an channel-spatial attention mechanism with variational few-shot prototype generation through the Bhattarcharyya metric, we present an inventive method that boosts feature extraction and conserves essential class knowledge pertinent to each task at hand. Our findings make a significant contribution towards improving comprehension of how cutting-edge techniques can work harmoniously together to address both benchmark experimental settings and intricate real-world issues including emergency response situations brought about by natural disasters. In all benchmark data sets, our approach outperformed the SOTA FSLs in the AA scores in the 5-way-5-shot evaluation, while for the UAV datasets, our approach outperformed the corresponding SOTA in the AA and macro-f1 scores in the 5-way-1-shot evaluation in general, demonstrating the feasibility and effectiveness of our FSL design in allowing better classification of a wide variety of image-based disaster scenery. Future works will investigate the extension of ABHFA-Net towards end-to-end real-time disaster perception by directly exploiting the pixel-level semantic annotations through joint few-shot classification and semantic segmentation. Additional directions include improving robustness under more challenging environmental degradations, deployment on embedded edge AI platforms for onboard UAV inference, and incorporating advanced data augmentation and generative techniques, such as GAN-based augmentation \cite{dam2020mixture, dam2022latent}, to further enhance generalization under extremely limited supervision.


\begin{appendices}

\section{Derivation of the Modified ELBO ($ELBO^{*}$) for Bhattarcharyya Coefficient}\label{secA1}

In this appendix section, we provide a derivation for the modified ELBO ($ELBO^{*}$) that relates to the Bhattarcharyya distance $D_{BD}$ (and hence the coefficient $BC$). We start with the definition of $BC$ as

\begin{equation} \label{eqA1}
  BC = \int \sqrt{q_{\phi}(z|\mathcal{S})p_{\theta}(z|\mathcal{T}, \mathcal{S})}dz.
\end{equation}

The derivation works best if we consider the Bhattarcharryya distance $D_{BD}$ directly, which is defined as the negative of the natural logarithm of $BC$ as $D_{BD} = -ln(BC)$. From this we take the derivative with respect to $z$ on both sides to get

\begin{equation} \label{eqA2}
   \frac{d}{dz}\left(BC\right) = \sqrt{q_{\phi}(z|\mathcal{S})p_{\theta}(z|\mathcal{T}, \mathcal{S})}.
\end{equation}

Taking natural logarithm on both sides and expanding out the right-hand side, we get

\begin{equation} \label{eqA3}
   ln\left(\frac{d}{dz}\left(e^{-D_{BD}}\right)\right) = \frac{1}{2}ln(q_{\phi}(z|\mathcal{S})) + \frac{1}{2} ln(p_{\theta}(z|\mathcal{T}, \mathcal{S})).
\end{equation}

We now use the property that $\int q_{\phi}(z|\mathcal{S})dz = 1$ and introduced it on the left-hand side of the equation to get

\begin{equation} \label{eqA4}
   ln\left(\frac{d}{dz}\left(e^{-D_{BD}}\int q_{\phi}(z|\mathcal{S})dz\right)\right) = \frac{1}{2} ln(q_{\phi}(z|\mathcal{S})) + \frac{1}{2} ln(p_{\theta}(z|\mathcal{T}, \mathcal{S})),
\end{equation}


where $D_{BD}$ is explicitly independent on z, so we can bring the $e^{-D_{BD}}$ term out of the integral to obtain

\begin{equation} \label{eqA5}
   ln\left(e^{-D_{BD}}\frac{d}{dz}\int q_{\phi}(z|\mathcal{S}) dz \right) = \frac{1}{2} ln(q_{\phi}(z|\mathcal{S})) + \frac{1}{2} ln(p_{\theta}(z|\mathcal{T},\mathcal{S})),
\end{equation}

\begin{equation} \label{eqA6}
   ln\left(e^{-D_{BD}}q_{\phi}(z|\mathcal{S}) \right) = \frac{1}{2} ln(q_{\phi}(z|\mathcal{S})) + \frac{1}{2} ln(p_{\theta}(z|\mathcal{T},\mathcal{S})).
\end{equation}

Expanding out the left-hand side, taking note that $ln(e^{-D_{BD}})= -D_{BD}$,

\begin{equation} \label{eqA7}
   ln\left(q_{\phi}(z|\mathcal{S})\right)-D_{BD} = \frac{1}{2} ln(q_{\phi}(z|\mathcal{S})) + \frac{1}{2} ln(p_{\theta}(z|\mathcal{T}, \mathcal{S})), 
\end{equation}

and multiplying both sides by $q_{\phi}(z|\mathcal{S})$, followed by integrating both sides with respect to dz, we get

\begin{equation}
\begin{split} \label{eqA8}
   \int q_{\phi}(z|\mathcal{S}) ln\left(q_{\phi}(z|\mathcal{S})\right) dz - \int D_{BD} q_{\phi}(z|\mathcal{S}) dz = \\\frac{1}{2} \int q_{\phi}(z|\mathcal{S}) ln(q_{\phi}(z|\mathcal{S})) dz + \frac{1}{2} \int q_{\phi}(z|\mathcal{S}) ln(p_{\theta}(z|\mathcal{T}, \mathcal{S})) dz,
\end{split}
\end{equation}

\begin{equation}
\begin{split} \label{eqA9}
   \frac{1}{2}\int q_{\phi}(z|\mathcal{S}) ln\left(q_{\phi}(z|\mathcal{S})\right) dz - \int D_{BD} q_{\phi}(z|\mathcal{S}) dz =  \frac{1}{2} \int q_{\phi}(z|\mathcal{S}) ln(p_{\theta}(z|\mathcal{T}, \mathcal{S})) dz, \\
   \frac{1}{2}\int q_{\phi}(z|\mathcal{S}) ln\left(q_{\phi}(z|\mathcal{S})\right) dz - \frac{1}{2} \int q_{\phi}(z|\mathcal{S}) ln(p_{\theta}(z|\mathcal{T}, \mathcal{S})) dz = \int D_{BD} q_{\phi}(z|\mathcal{S})dz. \\
\end{split}
\end{equation}


Add $\frac{1}{2}\int q_{\phi}(z|\mathcal{S})ln(p(z,\mathcal{T}))dz - \frac{1}{2}\int q_{\phi}(z|\mathcal{S})ln(p(z,\mathcal{T})) dz$ to the left-hand side of \ref{eqA9}:



\begin{equation} 
\begin{split} \label{eqA10}
   \frac{1}{2}\int q_{\phi}(z|\mathcal{S})ln\left(\frac{q_{\phi}(z|\mathcal{S})}{p(z,\mathcal{T})}\right)dz - \frac{1}{2}\int q_{\phi}(z|\mathcal{S})ln\left(\frac{p(z,\mathcal{T})}{p_{\theta}(z|\mathcal{T}, \mathcal{S})}\right)dz = \int D_{BD} q_{\phi}(z|\mathcal{S})dz, \\
   -\frac{1}{2}\int q_{\phi}(z|\mathcal{S})ln\left(\frac{p(z,\mathcal{T})}{q_{\phi}(z|\mathcal{S})}\right)dz - \frac{1}{2}\int q_{\phi}(z|\mathcal{S})ln\left(\frac{p(z,\mathcal{T})}{p_{\theta}(z|\mathcal{T}, \mathcal{S})}\right)dz = \int D_{BD} q_{\phi}(z|\mathcal{S})dz.
\end{split}
\end{equation}



We defined our modified Evidence Lower Bound term ($ELBO'$) as

\begin{equation} \label{eqA11}
   ELBO' = \int q_{\phi}(z|\mathcal{S}) ln \left(\frac{p(z,\mathcal{T})}{q_{\phi}(z|\mathcal{S})}\right)dz
\end{equation}

which differs from the original ELBO in the usage of the natural logarithm term ($ln$) instead of log to the base 10 ($log$).



Hence \ref{eqA10} can be rewritten as


\begin{equation}
\begin{split} \label{eqA12}
  -\frac{1}{2}ELBO' - \frac{1}{2}\int q_{\phi}(z|\mathcal{S})ln\left(\frac{p(z,\mathcal{T})}{p_{\theta}(z|\mathcal{T}, \mathcal{S})}\right)dz = \int D_{BD} q_{\phi}(z|\mathcal{S})dz,\\ 
  \frac{1}{2}ELBO' + \frac{1}{2}\int q_{\phi}(z|\mathcal{S})ln\left(\frac{p(z,\mathcal{T})}{p_{\theta}(z|\mathcal{T}, \mathcal{S})}\right)dz = -\int D_{BD} q_{\phi}(z|\mathcal{S})dz.\\ 
\end{split}
\end{equation}

We made use of $\int q_{\phi}(z|\mathcal{S})dz = 1$ again, as well as the fact that $D_{BD}$ is explicitly independent of $z$ on the right hand side of \ref{eqA12} to arrive at

\begin{equation}
\begin{split} \label{eqA13}
  \frac{1}{2}ELBO' + \frac{1}{2}\int q_{\phi}(z|\mathcal{S})ln\left(\frac{p(z,\mathcal{T})}{p_{\theta}(z|\mathcal{T}, \mathcal{S})}\right)dz = -D_{BD},\\ 
  ELBO' + \int q_{\phi}(z|\mathcal{S})ln\left(\frac{p(z,\mathcal{T})}{p_{\theta}(z|\mathcal{T}, \mathcal{S})}\right)dz = -2D_{BD}.\\ 
\end{split}
\end{equation}




We expand out $p(z,\mathcal{T})$ using Bayes' rule as $p(z|\mathcal{T})p(z)$ and straightforwardly obtained the reconstruction term $\int q_{\phi}(z|\mathcal{S})ln(p(z|\mathcal{T}))dz$ as

\begin{equation}
\begin{split} \label{eqA14}
  ELBO' + \int q_{\phi}(z|\mathcal{S})ln\left(\frac{p(z)}{p_{\theta}(z|\mathcal{T}, \mathcal{S})}\right)dz + \int q_{\phi}(z|\mathcal{S})ln(p(z|\mathcal{T})) dz= -2D_{BD},\\ 
    \int q_{\phi}(z|\mathcal{S})ln(p(z|\mathcal{T})) dz + 2D_{BD} = -\int q_{\phi}(z|\mathcal{S})ln\left(\frac{p(z)}{p_{\theta}(z|\mathcal{T}, \mathcal{S})}\right)dz - ELBO'.\\ 
\end{split}
\end{equation}

Finally, we defined $ELBO^{*} = -\int q_{\phi}(z|\mathcal{S})ln\left(\frac{p(z)}{p_{\theta}(z|\mathcal{T}, \mathcal{S})}\right)dz) - ELBO'$ and we arrived at


\begin{equation}
\begin{split} \label{eqA15}
   ELBO^{*} =  \int q_{\phi}(z|\mathcal{S}) ln(p_{\theta}(z|\mathcal{T})) dz + 2D_{BD},
\end{split}
\end{equation}

which is equation \ref{eq4.5}. \\

\section{Convergence Analysis}\label{secA2}

In this appendix section we analyses some theoretical properties of the modified ELBO (derived in the previous section) and the Bhattarcharyya loss $\ell_{BHAS}$, including their convergence analysis and stability proof.

For VAE optimization, in which our ABHFA-Net is part of, the convergence guarantees is dependent on the convexity and smoothness of the parameter space. The Bhattarcharyya distance utilized in our VAE optimization is a jointly (and strictly) convex measure in the pair of distributions $p_{\theta}$, $p^{'}_{\theta}$ and $q_{\phi}$. This means that for any distributions ($p_{\theta}$, $p^{'}_{\theta}$, $q_{\phi}$, $q^{'}_{\phi}$)) and a mixing parameter $\lambda\in[0,1]$, the mathematical inequality as shown below will hold true:

\begin{equation}
\begin{split} \label{eqB1}
    D_{BD}\left(\lambda p_{\theta}(z|\mathcal{T}, \mathcal{S}) + (1-\lambda)p^{'}_{\theta}(z|\mathcal{T}, \mathcal{S}), \lambda q_{\phi}(z|\mathcal{S}) + (1-\lambda)q^{'}_{\phi}(z|\mathcal{S})\right)\leq \\
    \lambda D_{BD}\left(p_{\theta}(z|\mathcal{T}, \mathcal{S}), \lambda q_{\phi}(z|\mathcal{S})\right) + (1-\lambda)D_{BD}\left(p^{'}_{\theta}(z|\mathcal{T}, \mathcal{S}),q^{'}_{\phi}(z|\mathcal{S})\right).
\end{split}
\end{equation}

On the surface, the aforementioned jointly convexity property of $D_{BD}$ allows gradient descent-based optimization algorithm, such as our VAE (which additionally utilized the reparameterization trick to make the gradient descent possible) to converge to a global minima rather than getting trapped in local minimas when minimizing $D_{BD}$ with respect to the differing probability distributions. To further prove that the global minima found is indeed unique, we can show that the Hessian matrix of $D_{BD}$ is semi-definite within all possible domain. We first consider the discrete version of $D_{BD}$ 

\begin{equation} \label{eqB2}
    D_{BD} = -ln\sum^{n}_{i=1} \sqrt{p_{i}q_{i}}
\end{equation}

and consider the Bhattarcharyya coefficient as 

\begin{equation} \label{eqB3}
    S(p) = \sum^{n}_{i=1} \sqrt{p_{i}q_{i}} = \sum^{n}_{i=1} a_{i}\sqrt{p_{i}}
\end{equation}

where $a_{i} = \sqrt{q_{i}}$. Therefore $D_{BD}$ reads as $D_{BD} = -ln(S(p))$ and the first derivative can be derived as 

\begin{equation} \label{eqB4}
    \frac{\partial D_{BD}}{\partial p_{i}} = \frac{-\partial ln(S(p))}{\partial p_{i}} = \frac{-1}{S(p)}\frac{\partial S(p)}{\partial p_{i}} = \frac{-1}{S(p)}\frac{\partial \left(\sum^{n}_{i=1} a_{i}\sqrt{p_{i}}\right)}{\partial p_{i}} = \frac{-1}{2S(p)}\left(\frac{a_{i}}{\sqrt{p_{i}}}\right). 
\end{equation}

The second derivative can be split into two cases: (i) For $i\neq j$:

\begin{equation} \label{eqB5}
    \frac{\partial^{2}D_{BD}}{\partial p_{i}p_{j}} = \frac{\partial}{\partial p_{i}} \left(\frac{-1}{2S(p)}\left(\frac{a_{j}}{\sqrt{p_{j}}}\right)\right) = \frac{a_{i}a_{j}}{4S^{2}\sqrt{p_{i}p_{j}}},
\end{equation}

while (ii) for $i=j$:

\begin{equation} \label{eqB6}
    \frac{\partial^{2}D_{BD}}{\partial p_{i}^{2}} = \frac{a_{i}}{4Sp^{\frac{3}{2}}_{i}} + \frac{a^{2}_{i}}{4S^{2}p_{i}}.
\end{equation}

Therefore the Hessian $\nabla^{2}D_{BD} = \left(\frac{\partial^{2}D_{BD}}{\partial p_{i}^{2}} \frac{\partial^{2}D_{BD}}{\partial p_{j}^{2}} \right)-\frac{\partial^{2}D_{BD}}{\partial p_{i}p_{j}}$ reads 

\begin{equation} \label{eqB7}
    \nabla^{2}D_{BD} = \frac{1}{4S^{2}}\left(\frac{a_{i}}{\sqrt{p_{i}}}\right)\left(\frac{a_{i}}{\sqrt{p_{i}}}\right)^{T} + \frac{1}{4S}diag \left(\frac{a_{i}}{p^{\frac{3}{2}}_{i}}\right).
\end{equation}

Taking any n-dim vector $x \in \mathbb{R}^{n}$ and perform the quadratic form $x^{T}(\nabla^{2}D_{BD})x$ (as it yielded a scalar): 

\begin{equation} \label{eqB8}
    x^{T}(\nabla^{2}D_{BD})x = \frac{1}{4S^{2}}\left(\frac{a_{i}x^{T}}{\sqrt{p_{i}}}\right)^{2} + \frac{1}{4S}\sum^{n}_{i=1} \frac{a_{i}x^{2}_{i}}{p_{i}^{3/2}}.
\end{equation}

Since the Bhattarcharyya coefficient must be positive $S > 0$, $p_{i} \geq 0$, $a_{i} = \sqrt{q_{i}} \geq 0$ (as both $p_{i}$ and $a_{i}$ are probability distribution), we arrived at the conclusion that \\

\begin{equation} \label{eqB9}
    x^{T}(\nabla^{2}D_{BD})x \geq 0
\end{equation}

and thus $\nabla^{2}D_{BD} \geq 0$, implying that the corresponding Hessian Matrix is indeed semi-definite across all $\mathbb{R}$.

\section{Stability Analysis}\label{secA3}

In this appendix section we illustrate the stability analysis of the Bhattarcharyya distance, the modified ELBO ($ELBO^{*}$), and the BHAS Loss $\ell_{BHAS}$. \\


\textbf{1) Stability of $D_{BD}$}. Assuming that the probability distribution $p$, $q$ are perturbed by $\delta p$ and $\delta q$ respectively such that $p' = p + \delta p$ and $q' = q + \delta q$. The Bhattarcharyya coefficient thus satisfies the following Lipschitz stability conditions: 

\begin{equation}
\begin{split} \label{C1}
   |BC(p',q') - BC(p,q)| \leq L(||\delta p|| + ||\delta q||),
\end{split}
\end{equation}

for some finite constant L. Since $D_{BD} = -ln(BC)$, by the mean value theorem

\begin{equation}
\begin{split} \label{C2}
   |D_{BD}(p',q') - D_{BD}(p,q)| = |-ln (BC(p',q')) + ln (BC(p,q))| \\
   =\left|\frac{1}{\zeta}\right||BC(p',q') - BC(p,q)| \\
   \leq \frac{1}{BC_{min}}|BC(p',q') - BC(p,q)| \\
   \leq \frac{L}{BC_{min}}(||\delta p|| + ||\delta q||)
\end{split}
\end{equation}

provided that $BC_{min} > 0$. In line 2 of equation \ref{C2} we have made used of the property $|f(x_{2})-f(x_{1})| = |f'(\zeta)||x_{2}-x_{1}|$ (from $f(x_{2})-f(x_{1}) = f'(\zeta)(x_{2}-x_{1})$), where $f(x) = -ln(x)$, $x = BC$ (and so $f(x) = -ln(BC)$), and $\zeta \in [min(x_{1},x_{2}), max(x_{1},x_{2})]$. In line 3 we have utilized $|f'(\zeta)| = \frac{1}{\zeta}$ (since $f(x) = -\frac{1}{x}$), as well as the observation that $BC \geq BC_{min} > 0$. This implies that $\zeta \geq BC_{min}$ (as $\zeta$ cannot be 0) and thus their reciprocal must also be true (albeit with sign reversal). In line 4 we have utilized the Lipschitz stability condition \ref{C1}. Since equation \ref{C2} is true, $D_{BD}$ must be locally Lipschitz continous, and any small perturbations in the latent distributions induced only bounded perturbation, making $D_{BD}$ stable.\\

\textbf{2) Stability of $ELBO^{*}$}. We recalled the relationship between $ELBO^{*}$ and $D_{BD}$

\begin{equation}
\begin{split} \label{C3}
   ELBO^{*} = \int q_{\phi}(z|\mathcal{S})ln\left(p_{\theta}(z|\mathcal{T})\right) dz + 2D_{BD},
\end{split}
\end{equation}

which implied the reconstruction term as $\int q_{\phi}(z|\mathcal{S})ln\left(p_{\theta}(z|\mathcal{T})\right)$ will be finite whenever $p_{\theta}(z|\mathcal{T}) > 0$.  Since $D_{BD}$ is bounded in the lower end by 0 (i.e., $D_{BD}\geq 0$) and hence is finite, the addition of the two finite terms remains finite, instead of diverging to infinity. \\

\textbf{3) Stability of BHAS Loss $\ell_{BHAS}$}. We recall the mathematical definition of $\ell_{BHAS}$, rewritten in terms of $D^{2}_{H}$ as

\begin{equation}  \label{C4}
   \ell_{BHAS} = -log\left(\frac{e^{\left(\frac{1 - D_{H}^{2}}{\tau}\right)}}{\sum_{i=1}^{2N} e^{\left(\frac{1 - D_{H}^{2}}{\tau}\right)}}\right).
\end{equation}

Since $0 \leq D^{2}_{H} \leq 1$, this implies that $0 \leq 1- D^{2}_{H} \leq 1$. Hence the exponential term appearing in both numerator and denominator of $\ell_{BHAS}$, $e^{\left(\frac{1 - D_{H}^{2}}{\tau}\right)}$ is bounded between 1 and $e^{\left(\frac{1}{\tau}\right)}$, i.e., $1 \leq e^{\left(\frac{1 - D_{H}^{2}}{\tau}\right)} \leq e^{\left(\frac{1}{\tau}\right)}$, provided that $\tau \neq 0$. Therefore $\ell_{BHAS}$ has bounded logits and hence prevent softmax value divergence, ensuring the stability of $\ell_{BHAS}$. \\

\section{Precision and Recall values for AIDER, CDD and MEDIC} \label{secA4}

\begin{table}[b!]
\centering
\caption{The precision and recall values along with their 95$\%$ confidence interval for 10 trials using the 5-way 1-shot and 5-way 5-shot learning evaluation on our AIDER subset simulation. The bolded values denote the best (highest) value obtained. \label{tabD1}}
\begin{tabular}{p{2.7cm}p{2cm}p{2cm}p{2cm}p{2cm}}
\hline
\textbf{Methods} & \textbf{Precision(1-shot)} & \textbf{Recall(1-shot)} & \textbf{Precision(5-shot)} & \textbf{Recall(5-shot)} \\
\hline
 Siamese Network & 65.3$\pm$0.18 & 65.3$\pm$0.15 & - & - \\
 Triplet Network & 68.3$\pm$0.66 & 68.3$\pm$0.72 & - & - \\
 ProtoNet & 58.7$\pm$0.55 & 58.7$\pm$0.59 & 77.8$\pm$0.31 & 77.8$\pm$0.24\\
 RelationNet & 48.3$\pm$0.57 & 48.3$\pm$0.57 & 79.4$\pm$0.13 & 79.4$\pm$0.12 \\
 MatchingNet & 41.5$\pm$0.26 & 41.5$\pm$0.27 & 76.1$\pm$1.00 & 76.1$\pm$1.06 \\
 SimpleShot & 59.4$\pm$0.58 & 59.4$\pm$0.58 & 76.5$\pm$0.16 & 76.5$\pm$0.22 \\
 BDCSPN & 49.9$\pm$0.67 & 49.9$\pm$0.67 & 65.6$\pm$0.22 & 65.6$\pm$0.24\\
 TIM & 62.9$\pm$1.73 & 62.9$\pm$1.71 & 71.2$\pm$1.13 & 71.2$\pm$1.16 \\
 Prototype FT & 61.2$\pm$0.45 & 61.2$\pm$0.44 & 68.7$\pm$0.74 & 68.7$\pm$0.75 \\
 Transductive FT & 60.8$\pm$1.0 & 60.8$\pm$1.04 & 67.4$\pm$1.10 &  67.4$\pm$1.13 \\
 VFA-Net & 67.5$\pm$0.43 & 67.5$\pm$0.45 & 75.5$\pm$0.46 & 75.5$\pm$0.46 \\
 Wasserstein VI & 64.4$\pm$0.96 & 64.4$\pm$0.96 & 69.6$\pm$1.04 & 69.6$\pm$1.04 \\
 HELA-VFA & 70.5$\pm$0.62 & 70.5$\pm$0.62 & 78.8$\pm$0.91 & 78.8$\pm$0.90 \\
 CVAE & 70.2$\pm$0.56 & 70.2$\pm$0.56 & 77.6$\pm$0.87 & 77.6$\pm$0.89 \\
 FSDM & 71.9$\pm$0.40 & 71.9$\pm$0.42 & 81.8$\pm$0.52 & 81.8$\pm$0.52 \\
 GL-ViT & 70.8$\pm$0.75 & 70.8$\pm$0.73 & 80.1$\pm$0.51 & 80.1$\pm$0.54 \\
 HC-Transformer & 71.8$\pm$0.44 & 71.8$\pm$0.47 & \textbf{82.5$\pm$0.81} & \textbf{82.5$\pm$0.82} \\
 SSL-ViT-16 & 71.1$\pm$0.71 & 71.1$\pm$0.72 & 80.7$\pm$0.75 & 80.7$\pm$0.74  \\
 \textbf{ABHFA-Net} & \textbf{72.2$\pm$1.05} & \textbf{72.2$\pm$1.05} & 81.4$\pm$0.67 & 81.4$\pm$0.67 \\
 \hline
\end{tabular}
\label{tab:precision_recall_results_AIDER}
\end{table}

\begin{table}
\centering
\caption{The precision and recall values along with their 95$\%$ confidence interval for 10 trials using the 5-way 1-shot and 5-way 5-shot learning evaluation on our CDD subset simulation. The bolded values denote the best (highest) value obtained.}\label{tabD2} \begin{tabular}{p{3cm}p{2cm}p{2cm}p{2cm}p{2cm}}
\hline
\textbf{Methods} & \textbf{Precision(1-shot)} & \textbf{Recall(1-shot)} & \textbf{Precision(5-shot)} & \textbf{Recall(5-shot)} \\
\hline
  Siamese Network & 63.2$\pm$0.61 & 63.2$\pm$0.59 & - & - \\
 Triplet Network & 64.5$\pm$0.55 & 64.6$\pm$0.54 & - & - \\
 ProtoNet & 57.7$\pm$1.10 & 57.6$\pm$1.13 & 70.3$\pm$1.05 & 70.4$\pm$1.07 \\
 RelationNet & 48.2$\pm$0.59 & 48.3$\pm$0.58 & 75.2$\pm$0.13 & 75.2$\pm$0.14 \\
 MatchingNet & 41.5$\pm$0.26 & 41.4$\pm$0.28 & 76.1$\pm$1.04 & 76.1$\pm$1.06 \\
 SimpleShot & 43.9$\pm$1.42 & 44.0$\pm$1.40 & 56.8$\pm$0.64 & 56.8$\pm$0.65 \\
 BDCSPN & 42.1$\pm$1.11 & 42.1$\pm$1.11 & 56.0$\pm$0.48 & 55.8$\pm$0.47 \\
 TIM & 59.2$\pm$0.49 & 59.2$\pm$0.52 & 67.3$\pm$0.87 & 67.4$\pm$0.90 \\
 Prototype FT & 58.3$\pm$0.88 & 58.5$\pm$0.92 & 68.1$\pm$0.45 & 68.0$\pm$0.45 \\
 Transductive FT & 55.0$\pm$0.56 & 55.1$\pm$0.58 & 67.9$\pm$0.41 & 67.9$\pm$0.41 \\
 VFA-Net & 64.2$\pm$0.31 & 64.2$\pm$0.32 & 72.2$\pm$0.46 & 72.2$\pm$0.46 \\
 Wasserstein VI & 64.5$\pm$1.05 & 64.5$\pm$1.05 & 68.4$\pm$1.21 & 68.4$\pm$1.21 \\
 HELA-VFA & 66.8$\pm$0.42 & 66.7$\pm$0.41 & 75.8$\pm$0.54 & 75.7$\pm$0.55 \\
 CVAE & 65.6$\pm$0.82 & 65.6$\pm$0.84 & 73.6$\pm$0.81 & 73.6$\pm$0.80 \\
 FSDM & 66.5$\pm$0.62 & 66.5$\pm$0.62 & 77.3$\pm$0.91 & 77.2$\pm$0.92 \\
 GL-ViT & 64.6$\pm$0.95 & 64.6$\pm$0.92 & 75.5$\pm$0.86 & 75.5$\pm$0.88\\
 HC-Transformer & 66.9$\pm$0.88 & 67.0$\pm$0.89 & \textbf{79.1$\pm$1.27} & \textbf{79.2$\pm$1.26} \\
 SSL-ViT-16 & 65.5$\pm$0.99 & 65.5$\pm$0.99 & 76.1$\pm$1.12 & 76.1$\pm$1.12  \\
 \textbf{ABHFA-Net} & \textbf{67.0$\pm$1.13} & \textbf{67.1$\pm$1.14} & 77.5$\pm$0.75 & 77.6$\pm$0.73 \\
 \hline 
\end{tabular}
\label{tab:precision_recall_results_CDD}
\end{table}

\begin{table}
\caption{The precision and recall values along with their 95$\%$ confidence interval for 10 trials using the 5-way 1-shot and 5-way 5-shot learning evaluation on our MEDIC subset simulation. The bolded values denote the best (highest) value obtained.} \label{tabD3}
\begin{tabular}{p{3cm}p{2cm}p{2cm}p{2cm}p{2cm}}
\hline
\textbf{Methods} & \textbf{Precision(1-shot)} & \textbf{Recall(1-shot)} & \textbf{Precision(5-shot)} & \textbf{Recall(5-shot)} \\
\hline
 Siamese Network & 60.5$\pm$1.55 & 60.4$\pm$1.57 & - & - \\
 Triplet Network & 58.3$\pm$1.54 & 58.4$\pm$1.56 & - & - \\
 ProtoNet & 37.1$\pm$1.36 & 37.3$\pm$1.40 & 57.8$\pm$1.55 & 57.9$\pm$1.59 \\
 RelationNet & 28.7$\pm$1.31 & 28.7$\pm$1.31 & 53.3$\pm$1.40 & 53.3$\pm$1.40\\
 MatchingNet & 40.4$\pm$1.03 & 40.5$\pm$1.05 & 48.9$\pm$1.21 & 48.9$\pm$1.20 \\
 SimpleShot & 46.4$\pm$0.84 & 46.3$\pm$0.86 & 57.7$\pm$0.57 & 57.8$\pm$0.56 \\
 BDCSPN & 43.6$\pm$0.81 & 43.6$\pm$0.80 & 61.7$\pm$0.56 & 61.7$\pm$0.56\\
 TIM & 54.3$\pm$0.69 & 54.5$\pm$0.71 & 63.7$\pm$1.25 & 63.8$\pm$1.26 \\
 Prototype FT & 44.1$\pm$1.88 & 44.3$\pm$1.89 & 62.2$\pm$0.62 & 62.3$\pm$0.60 \\
 Transductive FT & 46.4$\pm$0.60 & 46.5$\pm$0.59 & 63.9$\pm$1.20 & 63.8$\pm$1.21  \\
 VFA-Net & 55.0$\pm$1.00 & 55.0$\pm$0.99 & 63.5$\pm$1.38 & 63.5$\pm$1.38 \\
 Wasserstein VI & 57.0$\pm$1.01 & 57.0$\pm$1.02 & 62.9$\pm$1.43 & 62.9$\pm$1.44 \\
 HELA-VFA & 60.9$\pm$1.22 & 60.9$\pm$1.21 & 68.2$\pm$1.83 & 68.2$\pm$1.82 \\
 CVAE & 60.5$\pm$1.22 & 60.6$\pm$1.24 & 66.4$\pm$1.39 & 66.4$\pm$1.39 \\
 FSDM & 60.5$\pm$1.18 & 60.5$\pm$1.16 & \textbf{70.5$\pm$1.21} & \textbf{70.5$\pm$1.22}\\
 GL-ViT & 60.3$\pm$1.30 & 60.3$\pm$1.31 & 65.6$\pm$1.05 & 65.6$\pm$1.05\\
 HC-Transformer & 62.2$\pm$0.99 & \textbf{62.2$\pm$0.99} & 68.4$\pm$0.75 & 68.5$\pm$0.75 \\
 SSL-ViT-16 & \textbf{61.9$\pm$1.08} & 62.0$\pm$1.09 & 66.4$\pm$1.35 &  66.4$\pm$1.34  \\
 \textbf{ABHFA-Net} & \textbf{61.9$\pm$1.16} & 61.9$\pm$1.15 & 68.7$\pm$1.13 & 68.7$\pm$1.11\\
 \hline
\end{tabular}
\label{tab:precision_recall_results_MEDIC}
\end{table}

Table \ref{tab:precision_recall_results_AIDER}, \ref{tab:precision_recall_results_CDD} and \ref{tab:precision_recall_results_MEDIC} depicts the precision and recall values of the selected approaches with our ABHFA-Net for the 1-shot and 5-shot evaluation paradigm, along with their corresponding 95$\%$ confidence interval for 10 trials in each table. 

Interestingly, the precision and recall values are almost the same as that of the F1-scores for all shot configuration in all the models throughout the datasets. The close agreement among accuracy, macro precision, macro recall, and macro F1 is expected because the few-shot evaluation episodes has become class-balanced after under-sampling, with the same number of query samples drawn from each class. This indicates that the baseline errors are relatively evenly distributed across classes rather than being dominated by a single majority class. Hence the reported accuracy remains a representative summary metric, while precision, recall, and F1 provide additional confirmation of class-balanced behavior. \\

\end{appendices}

\bmhead{Acknowledgment}

This research/project is supported by the Civil Aviation Authority of Singapore and NTU under their collaboration in the Air Traffic Management Research Institute. Any opinions, findings and conclusions or recommendations expressed in this material are those of the author(s) and do not necessarily reflect the views of the Civil Aviation Authority of Singapore.

\section*{Declarations}

\begin{itemize}
\item \textbf{Funding Information}: This research is funded under the ATM-CAAS Leaders' Track Scholarship, under the NTU ATMRI-CAAS collaboration.
\item \textbf{Competing interests}: On behalf of all authors, the corresponding author states that there is no conflict of interest.
\item \textbf{Ethical Approval for Human and /or Animals Research}: Not applicable.
\item \textbf{Availability of Data and Materials}: All data will be made available upon request.
\item \textbf{Author contributions (CreDiT Authorship Statement)}: \textbf{Gao Yu Lee}: Conceptualization, Investigation, Methodology, Software, Writing - original draft, Writing- review $\&$ editing. \textbf{Tanmoy Dam}:  Conceptualization, Investigation, Supervision, Writing- review $\&$ editing. \textbf{Md Mefahul Ferdaus}: Supervision, Writing- review $\&$ editing. \textbf{Daniel Puiu Poenar}: Supervision. \textbf{Vu N. Duong}: Supervision.
\end{itemize}

\section*{Conflict of Interest}

On behalf of all authors, the corresponding author states that there is \textbf{no conflict of interest.}

\bibliography{ref}

\end{document}